\documentclass{ecai} 
\usepackage{latexsym}
\usepackage{amssymb}
\usepackage{amsmath}
\usepackage{amsthm}
\usepackage{booktabs}
\usepackage{enumitem}
\usepackage{graphicx}
\usepackage{color}
\usepackage[colorlinks = true,
            linkcolor = blue,
            urlcolor  = blue,
            citecolor = blue,
            anchorcolor = blue]{hyperref}

\usepackage{algorithm}
\usepackage{algpseudocode}
\usepackage{algcompatible}

\usepackage{xcolor, colortbl}
\definecolor{LightCyan}{rgb}{0.7,0.7,1}
\definecolor{LightLightGray}{gray}{0.9}

\usepackage{multirow}
\usepackage{caption}
\usepackage{subcaption}

\newcommand{\BibTeX}{B\kern-.05em{\sc i\kern-.025em b}\kern-.08em\TeX}

\makeatletter
\def\blfootnote{\xdef\@thefnmark{}\@footnotetext}
\makeatother
\begin{document}

%%%%%%%%%%%%%%%%%%%%%%%%%%%%%%%%%%%%%%%%%%%%%%%%%%%%%%%%%%%%%%%%%%%%%%%%

\begin{frontmatter}

%%% Use this command to specify your submission number.
%%% In doubleblind mode, it will be printed on the first page.

\paperid{123} 

%%% Use this command to specify the title of your paper.

\title{Nonlinear Concept Erasure:\\a Density Matching Approach}

%%% Use this combinations of commands to specify all authors of your 
%%% paper. Use \fnms{} and \snm{} to indicate everyone's first names 
%%% and surname. This will help the publisher with indexing the 
%%% proceedings. Please use a reasonable approximation in case your 
%%% name does not neatly split into "first names" and "surname".
%%% Specifying your ORCID digital identifier is optional. 
%%% Use the \thanks{} command to indicate one or more corresponding 
%%% authors and their email address(es). If so desired, you can specify
%%% author contributions using the \footnote{} command.

\author[A]{\fnms{Antoine}~\snm{Saillenfest}\thanks{Corresponding Author. Email: a.saillenfest@groupeonepoint.com.}}
\author[A]{\fnms{Pirmin}~\snm{Lemberger}\thanks{Corresponding Author. Email: p.lemberger@groupeonepoint.com.}}

\address[A]{onepoint, 29 rue des Sablons, 75116 Paris (France)}

%%% Use this environment to include an abstract of your paper.

\begin{abstract}
Ensuring that neural models used in real-world applications cannot infer sensitive information, such as demographic attributes like gender or race, from text representations is a critical challenge when fairness is a concern. We address this issue through concept erasure, a process that removes information related to a specific concept from distributed representations while preserving as much of the remaining semantic information as possible. Our approach involves learning an orthogonal projection in the embedding space, designed to make the class-conditional feature distributions of the discrete concept to erase indistinguishable after projection. By adjusting the rank of the projector, we control the extent of information removal, while its orthogonality ensures strict preservation of the local structure of the embeddings. Our method, termed $\overline{\mathrm{L}}$EOPARD, achieves state-of-the-art performance in nonlinear erasure of a discrete attribute on classic natural language processing benchmarks. Furthermore, we demonstrate that $\overline{\mathrm{L}}$EOPARD effectively mitigates bias in deep nonlinear classifiers, thereby promoting fairness.
\end{abstract}

\end{frontmatter}

%%%%%%%%%%%%%%%%%%%%%%%%%%%%%%%%%%%%%%%%%%%%%%%%%%%%%%%%%%%%%%%%%%%%%%%%

\section{Introduction}
\label{introduction}

Pre-trained distributed representations, or embeddings, capture similarity or semantic meaning and are widely used in numerous machine learning applications. In certain critical scenarios, preventing a system from encoding specific concepts poses a significant challenge. A substantial body of research in fairness focuses on learning representations that exclude sensitive concepts, such as demographic attributes like gender or race \cite{ravfogel2022adversarial, belrose2023leace, basu2022learning}. Similarly, in explainability research, removing a concept from a model's internal representation enables estimating its causal effect on a given task \cite{feder2021causalm, abraham2022cebab, lemberger2024explaining}. Consequently, there is growing interest in developing methods for learning representations that disentangle underlying concepts.\blfootnote{\newline\noindent\textit{Preprint (incl. Supplementary material)\newline To appear in Proceedings of ECAI 2025 - 28th European Conference on AI}}

\begin{figure}[t!]
    \centering
    \includegraphics[width=0.8\linewidth]{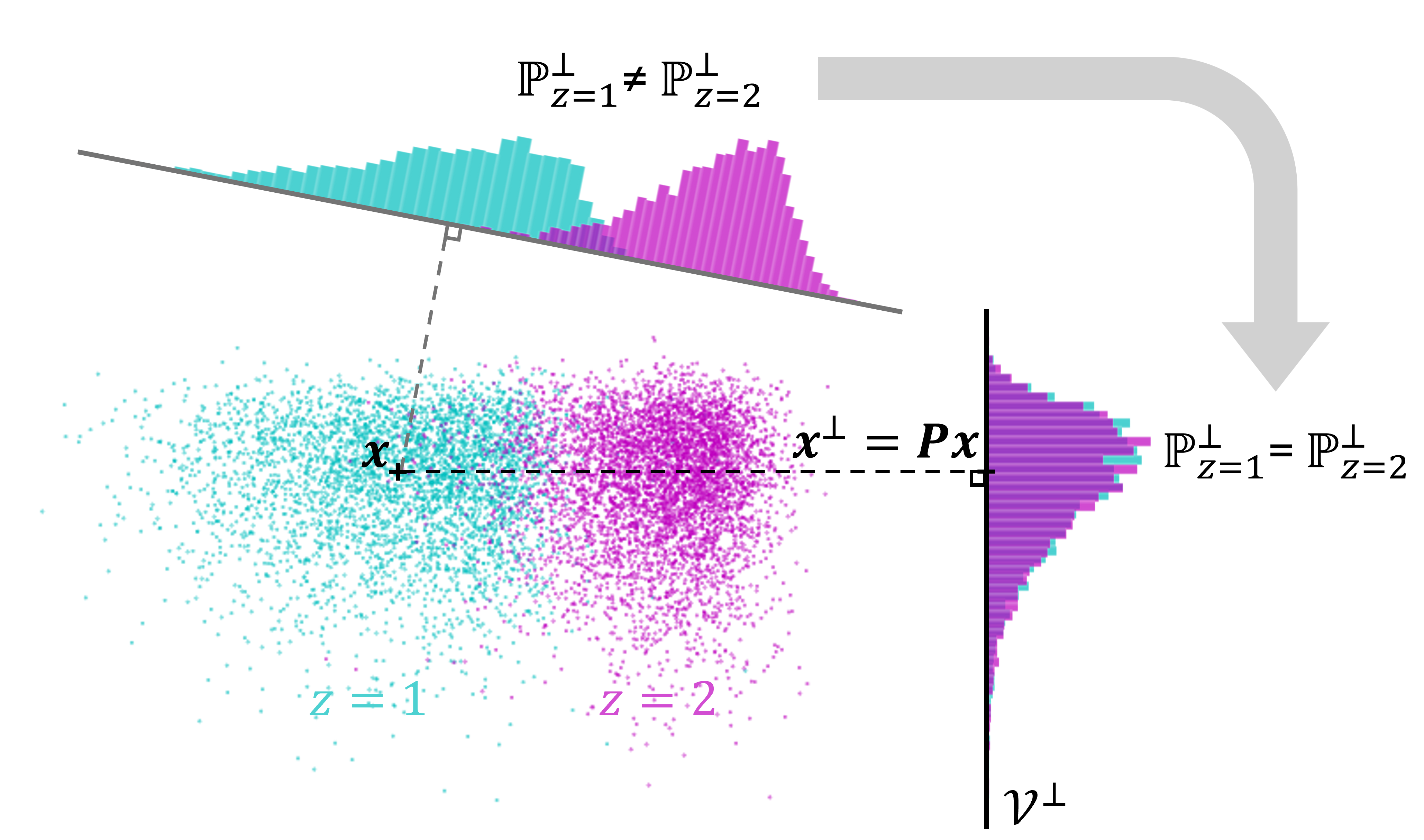}
    \caption{$\overline{\mathrm{L}}$EOPARD erases a discrete attribute $z$ by learning an orthogonal projection $P$ onto a subspace $\mathcal{V}^\perp$ of the feature space in which the distributions of the projected samples $x^\perp = Px$ are indistinguishable across values of the protected attribute to erase $z$.}
    \label{fig:LEOPARD-main-fig}
    \par\bigskip
    \par\bigskip
\end{figure}

In this paper, we address the task of nonlinear concept erasure \cite{ravfogel2022linear}, which involves balancing two competing objectives: first removing concept-related information from a set of representations to ensure they are no longer predictive of the concept after erasure, and second, preserving as much information as possible from the original representation space. Nonlinear erasure refers to the impossibility for a nonlinear predictor, such as an MLP, to recover the concept after erasure. Our focus is on erasing discrete concepts, including demographic attributes such as gender or race. Specifically, we investigate post-hoc concept erasure, where representations are fixed and pre-trained (e.g., GloVe \cite{pennington2014glove} or BERT embeddings \cite{devlin2019BERT} in NLP). Importantly, we tackle concept erasure in an unconstrained setting, which means that post-erasure representations are not optimized for any predefined downstream task.

We propose a simple yet novel concept erasure method which learns an orthogonal projection in the embedding space. The optimization process is driven by a density-matching erasure objective, ensuring that the class-conditional distributions of the representations after erasure are indistinguishable. The projection rank can be adjusted to control the extent of information removal. Leveraging orthogonal projections satisfies the dual objective of achieving non-invertible erasure while preserving the local structure of the feature space. We refer to our framework as $\overline{\mathrm{\textbf{L}}}$\textbf{EOPARD}: \textbf{Non}-\textbf{L}inear concept \textbf{E}rasure via \textbf{O}rthogonal \textbf{P}rojection and \textbf{A}lignment of post-erasure \textbf{R}epresentation \textbf{D}istributions. To evaluate the effectiveness of our approach, we conducted experiments demonstrating state-of-the-art performance in discrete concept erasure for the task of demographic attribute removal on standard NLP benchmarks. Furthermore, we show empirically that $\overline{\mathrm{L}}$EOPARD can improve fairness of downstream classifiers with moderate impact on their performance.
Our contributions are:
\begin{itemize}
    \item We introduce $\overline{\mathrm{L}}$EOPARD, a novel method for unconstrained nonlinear concept erasure relying on orthogonal projections and a class-conditional density matching objective. 
    \item We demonstrate that $\overline{\mathrm{L}}$EOPARD achieves state-of-the-art performance on many NLP benchmarks for the task of demographic attributes erasure. 
    \item We demonstrate the utility of $\overline{\mathrm{L}}$EOPARD in improving the fairness of deep classifiers with a moderate impact on performance.
\end{itemize}

\section{Related work}
\label{related_work}

Linear concept erasure is a restriction of concept erasure where adversaries are assumed to use linear predictors only to recover the concept after erasure. Some approaches use adversarial methods to learn projections that erase concepts \cite{ravfogel-etal-2020-null, ravfogel2022linear}. More recently, closed-form solutions for linear concept erasure have been proposed, eliminating altogether the need for learning an erasure function  \cite{ravfogel2022linear, belrose2023leace}. These projections remove an optimal number of dimensions solely determined by the number of concept classes, which is usually small. High rank orthogonal projection naturally preserve the local structure of the embeddings. However, these methods struggle to fully remove concept-related information, as nonlinear predictors can still recover the concept after erasure \cite{ravfogel2022adversarial}. Our method also involves learning projections but addresses the general problem of nonlinear concept erasure.

Nonlinear concept erasure remains largely an open problem. Recent efforts can be grouped into three main approaches. The first two are adversarial approaches relying on a gradient reversal layer during training \cite{ganin2016domain, feder2021causalm}, and kernelized extensions of linear erasure methods \cite{elazar2018adversarial, ravfogel2022adversarial}. Both fall short of fully removing the protected information \cite{ravfogel2022adversarial}. The last one covers recent methods that have drawn on rate-distortion theory to achieve nonlinear erasure by dispersing embeddings within the same concept class, effectively decorrelating them \cite{basu2022learning, basu2024robust}. These two works however raise fundamental issues. First, relying on nonlinear erasure functions may be detrimental to concept-unrelated information preservation. Second, these works address concept erasure, which is non-invertible, by learning invertible functions, which only become nearly non-invertible after optimization. In this paper, we focus on learning erasure projector which are linear and non-invertible by design.

Constrained de-biasing, or erasure, refers to optimizing post-erasure representations for a predefined downstream task, such as a classification \cite{basu2022learning}, or a more generic one, such as Masked Language Modeling (MLM) for language models fine-tuning \cite{devlin2019BERT, feder2021causalm}. In contrast, downstream tasks are assumed unknown a priori in an unconstrained mode, making it essential to preserve semantic information for broad applicability. Semantic preservation is typically evaluated via semantic similarity tests (e.g., WordSim-353 \cite{agirre2009study}, WHEAT \cite{caliskan2017semantics}) or the preservation of the neighborhood structure \cite{basu2024robust}. This paper is a contribution to unconstrained erasure.

Another relevant line of research focuses on developing predictors for downstream tasks that are robust to interventions on protected variables. To ensure such invariance, Veitch et al.~(\citeyear{veitch2021counterfactual}), followed by subsequent works~\cite{makar2022causally}, propose a regularization procedure which depends on an underlying causal model. This model assumes the existence of a part of the input representation that is independent of the protected attribute. By penalizing discrepancies between relevant conditional output distributions, they enforce specific independence properties between the target variable and the predictive features. By contrast, our method modifies the input representations. We operate under the assumption that the target variable is not available a priori, and our objective is to extract a part of the embeddings that is independent of the protected concept.

Concept erasure is relevant to numerous applications, including bias mitigation for fairness \cite{belrose2023leace, feder2021causalm}, causal analysis for explainability \cite{abraham2022cebab, lemberger2024explaining}, and semantic analysis \cite{ravfogel2022linear}. End-to-end training is often limited by computational constraints or inaccessible internal model parameters. As a result fixed, pre-trained representations remain widely used for their versatility across domains. Most existing approaches are evaluated on text embeddings from pre-trained models like \textsc{GloVe} \cite{pennington2014glove}, BERT \cite{devlin2019BERT}, or GPT \cite{brown2020language} or on image representations \cite{ravfogel2022linear, basu2024robust}. The simplest setting for erasure involves categorical concepts, which encompass a wide range of applications, such as the removal of demographic biases related to gender or race in text embeddings \cite{bolukbasi2016man, dearteaga2019bias, basu2022learning, basu2024robust}. This paper focuses on \textit{post-hoc} discrete concept erasure, assuming fixed and pre-trained embeddings.

\section{Learning Projections with a Density Matching Objective} 

This section presents our optimization method, $\overline{\mathrm{L}}$EOPARD, which aims to achieve nonlinear erasure of a discrete concept from a set of representations through orthogonal projections. We first introduce a parametrization for learning an orthogonal projection of prescribed rank, and subsequently introduce a density matching criterion to select the projection that best removes the concept information.

Let $\mathcal{D}=\{ x_i\}_{i=1}^N$ denote a dataset of representations in $\mathbb{R}^d$ sampled i.i.d. from an underlying distribution $\mathbb{P}$. Each representation $x_i$ is associated with a discrete class label $z_i \in \mathcal{Z} := \{1, \dots, K \}$ corresponding to the concept to be erased. Let $\mathcal{C}=\{ z_i\}_{i=1}^N$. We define the class-specific subset $\mathcal{D}_{z=j} = \{x_i \in \mathcal{D} | z_i = j\}$, so that $\mathcal{D} = \bigcup_{j=1}^K \mathcal{D}_{z=j}$, and let $n_{z=j} = |\mathcal{D}_{z=j}|$ denote the cardinality of each subset. We denote by $\mathbb{P}_{z=j} := \mathbb{P}(.|z=j)$ the class-conditional distribution of samples.

\subsection{Learning Orthogonal Projections}
\label{sec:orthogonal_projections} 

Our approach follows the linear bias subspace hypothesis, which posits that, for a given set of representations, the information related to a specific concept is entirely contained within a subspace of the embedding space \cite{bolukbasi2016man, vargas2020exploring}. Consequently, we seek an orthogonal decomposition of the embedding space $\mathbb{R}^d$ as $\mathbb{R}^d = \mathcal{V}^\perp \oplus \mathcal{V}^\parallel$ where $\mathcal{V}^\perp$ is the subspace of dimension $r$ that is to be preserved, and $\mathcal{V}^\parallel$ is its orthogonal complement of dimension $d-r$ that is to be neutralized.

Since concept erasure is inherently a non-invertible operation, we enforce the erasure map to be non-invertible as well. Additionally, for simplicity we constrain the erasure map to be linear. Specifically, to ensure that the semantic content of the embeddings is maintained after erasure, we aim to preserve the local geometry of the embeddings through erasure, i.e., the distances between embeddings. This constrains the erasure function to be an orthogonal projection onto $\mathcal{V}^\perp$. Such a projection is fully characterized by a symmetric, idempotent matrix $P \in \mathbb{R}^{d \times d}$ of rank $r$ satisfying $P^\top = P$ and $P^2 = P$. Let $x^\perp$ denote the post-erasure representation of a representation $x$. The erasure operation is given by:

\begin{equation}
\label{eq:projection_function}
    x^\perp = Px.
\end{equation}

Standard linear algebra states that any rank-$r$ orthogonal projection matrix $P$ admits the decomposition: 

\begin{equation}
\label{eq:model}
    P=UU^\top,
\end{equation}

\noindent where $U \in \mathbb{R}^{d \times r}$ is a matrix with orthonormal columns, i.e. \mbox{$U^\top U = I_r$}, where $I_r$ is the $r \times r$ identity matrix. 

In the context of an optimization process, the parametrization from Eq.~\ref{eq:model} allows us to learn $U$, and consequently $P$, under the orthonormality constraint. To encourage this constraint during training, we define a \textit{projection loss} to be minimized as:

\begin{equation}
\label{eq:loss_proj}
    \mathcal{L}_p(U) = \| U^\top U - I_r \|_\mathrm{F}^2,
\end{equation}
where $\| \cdot \|_\mathrm{F}$ denotes the Frobenius norm.

\subsection{Class-Conditional Density Matching}
\label{sec:density-matching}

To ensure that the learned projection effectively removes concept-specific information, we propose a criterion based on matching class-conditional distributions of the erased representations. Specifically, we aim for the post-erasure class-conditional distributions \mbox{$\mathbb{P}_{z=j}^\perp := \mathbb{P}_{z=j}(.^\perp$)} to be indistinguishable across protected attribute classes.

We first consider the binary case $\mathcal{Z} = \{1,2\}$. Let $\mathbb{P}_{z=1}^\perp$ and $\mathbb{P}_{z=2}^\perp$ denote the post-erasure distributions corresponding to $z = 1$ and $z = 2$, respectively. Our objective is to minimize a divergence measure $\mathrm{dist}(\mathbb{P}_{z=1}^\perp, \mathbb{P}_{z=2}^\perp)$ between these two distributions. To quantify this distance, we use the Maximum Mean Discrepancy (MMD), chosen for its practical estimation from finite samples \cite{gretton2012kernel}. MMD simply computes the distance between the means of the embeddings of the distributions in a reproducing kernel Hilbert space (RKHS) $\mathcal{H}$ associated with a positive-definite kernel $k : \mathbb{R}^d \times \mathbb{R}^d \to \mathbb{R}$. The unbiased empirical estimator of MMD$^2$ \cite{gretton2012kernel} is given by:
\begin{equation}
\label{eq:mmd_est}
\resizebox{\columnwidth}{!}{%
$
\begin{aligned}
\widehat{\mathrm{MMD}^2}(\mathbb{P}_{z=1}^\perp, \mathbb{P}_{z=2}^\perp)
&= \frac{1}{n_{z=1}(n_{z=1} - 1)} \sum_{\substack{x, x' \in \mathcal{D}_{z=1} \\ x \neq x'}} k(x^\perp, x'^\perp) \\
&\quad + \frac{1}{n_{z=2}(n_{z=2} - 1)} \sum_{\substack{x, x' \in \mathcal{D}_{z=2} \\ x \neq x'}} k(x^\perp, x'^\perp) \\
&\quad - \frac{2}{n_{z=1}n_{z=2}} \sum_{x \in \mathcal{D}_{z=1}} \sum_{x' \in \mathcal{D}_{z=2}} k(x^\perp, x'^\perp).
\end{aligned}
$
}
\end{equation}

A common choice of kernel is the Gaussian kernel, defined as $k(x, y) = \exp\left( -\frac{1}{2\sigma^2} \|x - y\|^2 \right)$, where $\sigma > 0$ is a bandwidth parameter. This kernel has a feature expansion that includes an infinite number of terms, thereby capturing all orders of statistics. In this case the MMD distance between the two distributions vanishes if and only if they coincide. 

In the context of an optimization procedure, the empirical estimator introduced above naturally induces an objective function, we refer to as the \emph{erasure loss}, which is to be minimized with respect to the parameters $U$ defining the erasure function:
\begin{equation}
\label{eq:mmd_loss_binary}
    \mathcal{L}_\mathrm{MMD}(U; \mathcal{D}, \mathcal{C}) := \widehat{\mathrm{MMD}^2}(\mathbb{P}_{z=1}^\perp, \mathbb{P}_{z=2}^\perp).
\end{equation}

In the general case of a concept that can take more than two values, i.e. $K \geq 2$, a natural generalization of the density matching objective in Eq.~\ref{eq:mmd_loss_binary} is to consider the sum of pairwise discrepancies:
\begin{equation}
\label{eq:mmd_loss_multi}
    \mathcal{L}_\mathrm{MMD}(U; \mathcal{D}, \mathcal{C}) := \sum_{\substack{i, j \in \mathcal{Z} \\ i < j}} \widehat{\mathrm{MMD}^2}(\mathbb{P}_{z=i}^\perp, \mathbb{P}_{z=j}^\perp).
\end{equation}

As a side note, let us observe that, as stated in Belrose et al.~(\citeyear{belrose2023leace}), one condition for linearly erasing a concept is that \mbox{$\mathbb{E}_{x \sim \mathbb{P}}[x^\perp | z = j] = \mathbb{E}_{x \sim \mathbb{P}}[x^\perp]$} for all concept classes $j$. In the context of steering, a problem closely related to erasure, Singh et al. (\citeyear{singh2024representation}) extend prior conditioning approaches to enforce the equality of mean and variance of class-conditional distributions, deriving a closed-form solution that corresponds to minimizing the Wasserstein distance under Gaussian assumptions. Our density matching objective can be viewed as a much stronger constraint on the class-conditional distributions which pushes them towards being indistinguishable.

\begin{algorithm}[t!]
   \caption{$\overline{\mathrm{L}}$EOPARD}
   \label{alg:erasure_procedure}
\begin{algorithmic}[1]
   \STATE \textbf{Input:} Dataset $\mathcal{D} = \{x_i\}_{i=1}^N$, concept labels $\mathcal{C} = \{z_i\}_{i=1}^N$, rank $r$, mixing coefficient $\gamma \in \mathbb{R}^+$, initial matrix $U^{(0)} \in \mathbb{R}^{d \times r}$, number of training epochs $N_e$
   \STATE \textbf{Output:} Rank-$r$ orthogonal projection matrix $P$
   \FOR{$i = 1$ \textbf{to} $N_e$}
      \STATE Compute gradient: $\nabla_U \left( \gamma \mathcal{L}_p(U) + \mathcal{L}_\mathrm{MMD}(U; \mathcal{D}, \mathcal{C}) \right)$
      \STATE Update $U$ via gradient descent
   \ENDFOR
   \STATE \textit{\% Project onto the closest orthogonal projector to $UU^\top$ in Frobenius norm}
   \STATE $U, \Lambda = \text{EigenDecomposition}(UU^\top)$
   \STATE $U_r \leftarrow U[:, 1{:}r]$
   \STATE $P \leftarrow U_r {U_r}^\top$
\end{algorithmic}
\end{algorithm}

\begin{algorithm}[h]
   \caption{Cascaded $\overline{\mathrm{L}}$EOPARD}
   \label{alg:cascaded_erasure_procedure}
\begin{algorithmic}[1]
   \STATE \textbf{Input:} Dataset $\mathcal{D} = \{x_i\}_{i=1}^N$, concept labels $\mathcal{C} =\{z_i\}_{i=1}^N$, rank $r$, mixing coefficient $\gamma \in \mathbb{R}^+$, initial matrix $U^{(0)} \in \mathbb{R}^{(d-K+1) \times r}$, number of training epochs $N_e$
   \STATE \textbf{Output:} Rank-$r$ orthogonal projection matrix $P$
   \STATE Compute the orthogonalized LEACE projection $P_L$
   \STATE Compute $U_L \in \mathbb{R}^{d \times (d-K+1)}$ such that $P_L = U_L U_L^\top$ and $U_L^\top U_L = I_{d-K+1}$
   \STATE Erase the concept via $\overline{\mathrm{L}}$EOPARD using Algorithm \ref{alg:erasure_procedure}:
   \STATEx $P' = \overline{\mathrm{L}}\text{EOPARD}(\{U_L^\top x_i\}_{i=1}^N, \mathcal{C}, r, \gamma, U^{(0)}, N_e)$
   \STATE Compose the final orthogonal projection: $P \leftarrow U_L P' U_L^\top$
   \STATE \textit{\%$P$ is a rank-r orthogonal projection: $P = P^\top$, $P^2 = P$ 
   \STATEx and $\mathrm{rank}(P) = \mathrm{rank}(P') = r$}
\end{algorithmic}
\end{algorithm}

\subsection{$\overline{\mathrm{\textbf{L}}}$EOPARD}
\label{sec:-LEOPARD}

The optimization procedure for $\overline{\mathrm{L}}$EOPARD is outlined in Algorithm~\ref{alg:erasure_procedure}. The goal is to learn a rank-$r$ orthogonal projection $P$ that erases a discrete concept nonlinearly, parameterized via a matrix $U \in \mathbb{R}^{d \times r}$ such that $P = UU^\top$, as in Eq.~\ref{eq:model}. The learning objective to minimize during training combines the \emph{projection loss}~$\mathcal{L}_p$ (Eq.~\ref{eq:loss_proj}) and the \emph{erasure loss} $\mathcal{L}_\mathrm{MMD}$ (Eq.~\ref{eq:mmd_loss_multi}) and is given by:
\begin{equation}
    \label{eq:total_loss}
    \mathcal{L}(U; \mathcal{D},  \mathcal{C}) = \gamma \mathcal{L}_p(U) + \mathcal{L}_\mathrm{MMD}(U; \mathcal{D}, \mathcal{C}),
\end{equation}

\noindent where the mixing coefficient $\gamma > 0$ balances the two objectives.

At convergence, the matrix $P = UU^\top$ is not guaranteed to be exactly idempotent due to approximate enforcement of orthogonality constraints. To ensure that the final output is a valid rank-$r$ orthogonal projector, we compute the closest orthogonal projection matrix in Frobenius norm to the approximate projection matrix $UU^\top$ at the end of the training loop. This is achieved by performing an eigendecomposition of $UU^\top$ and retaining the top $r$ eigenvectors to construct a semi-orthogonal matrix $U_r$, yielding the final orthogonal projection matrix $P \leftarrow U_r U_r^\top$.

The orthogonal projector $P$ obtained through this procedure preserves the local geometric structure (i.e. the pairwise distances), a desirable property in representation spaces. In contrast, the approximate projection $UU^\top$ at the end of training, while potentially achieving better concept erasure, lacks guaranteed idempotence. The extent to which the final orthogonal projection retains the erasure properties of the approximate solution depends on their proximity in Frobenius norm which is quantified precisely by the final value of the projection loss $\mathcal{L}_p(U)$ (see Supplementary Material~\ref{app:proof} for a proof).

The number of trainable parameters in $\overline{\mathrm{L}}$EOPARD is $d \cdot r$, significantly lower than that of existing nonlinear erasure methods such as FaRM~\cite{basu2022learning} and KRaM~\cite{basu2024robust}, which involve training multi-layer perceptrons for concept removal.

In $\overline{\mathrm{L}}$EOPARD, the adjustable rank $r$ governs the trade-off between the preservation of information and the effectiveness of concept removal. There does not exist a universally optimal value for $r$; the rank adjustability constitutes a key strength of our method, enabling trade-offs between erasure and information preservation that are not possible with more rigid approaches.

\paragraph{Cascaded $\overline{\mathrm{\textbf{L}}}$EOPARD}

The erasure objective introduced in Section~\ref{sec:density-matching} minimizes the $\mathrm{MMD}$ distance between class-conditional distributions. With a characteristic kernel (as in \ref{sec:density-matching}), this encourages alignment of all moments. However, MMD-based matching does not explicitly enforce mean alignment, which is essential for effective linear erasure \cite{belrose2023leace}. As a result, the learned projection may fail to fully remove the targeted concept in a linear sense. To address this, we incorporate recent advances in concept erasure that ensure linear removal by construction, with negligible computational overhead.

\emph{Cascaded erasure} is a two-stage process that integrates linear and nonlinear erasure. In the first stage, we apply an orthogonalized LEACE projection~\cite{belrose2023leace} to linearly remove the concept. The embeddings are then projected onto the $(d - K + 1)$-dimensional image subspace of this projection. In the second stage, these linearly "sanitized" embeddings are further processed using a nonlinear erasure method. When instantiated with $\overline{\mathrm{L}}$EOPARD, it amounts to computing two orthogonal projections acting on nested subspaces. The overall cascaded projection remains orthogonal and inherits the benefits of both linear and nonlinear concept erasure. The full procedure for Cascaded $\overline{\mathrm{L}}$EOPARD is outlined in Algorithm~\ref{alg:cascaded_erasure_procedure}.

\section{Experimental Setup}

To evaluate our erasure method, we focus on the task of discrete concept erasure in text embeddings and its application to training fair classifiers. We compare our approach with SOTA methods for unconstrained nonlinear concept erasure FaRM \cite{basu2022learning} and KRaM \cite{basu2024robust}.

Code and data : \href{https://github.com/toinesayan/non-LEOPARD}{https://github.com/toinesayan/non-LEOPARD}.

\subsection{Datasets}

\paragraph{\textsc{GloVe}} is a dataset derived from the \textsc{GloVe} embeddings of the 150k most frequent words \cite{pennington2014glove}. It includes a selection of words categorized as male-biased, female-biased, and neutral based on the magnitude of their projection onto the gender direction, the principal component of the vector space formed by gendered word-pair differences, treating gender as ternary. This dataset, used in \cite{basu2024robust}, consists of 21,996 word embeddings stratified by gender, split 49\%/21\%/30\% across training, validation, and test.
% : 10,777 training samples (49\%), 4,620 validation samples (21\%), and 6,599 test samples (30\%).

\paragraph{\textsc{Bias In Bios}} is a real world dataset suited for studying gender biases in biography classification tasks \cite{dearteaga2019bias}. It consists in biographies collected through web scraping and labeled with binary gender and occupation (28 in total). We use the version from \cite{ravfogel-etal-2020-null}, which contains $\sim$98\% of the original dataset, as the full version is no longer publicly available. It consists of 399,423 biographies stratified by occupation, split 65\%/10\%/25\% across training, validation, and test.
% with 255,710 training samples (65\%), 39,369 validation samples (10\%), and 98,344 test samples (25\%).

\paragraph{\textsc{DIAL}} is a Twitter-based sentiment classification dataset constructed from the DeepMoji corpus~\cite{blodgett2016demographic}. Each tweet is annotated with a binary sentiment label (\textit{happy} or \textit{sad}) and a "race" label, which corresponds to the linguistic style of the tweet (\textit{African-American English} (AAE) or \textit{Standard American English} (SAE)). We use the version of the dataset made available by~\cite{basu2024robust}, which is balanced with respect to both race and sentiment labels. \textsc{DIAL} consists of 176k samples, split 91\%/4.5\%/4.5\% across training, validation, and test.
% 160{,}000 training samples (90.1\%), 8{,}000 validation samples (4.5\%), and 7{,}996 test samples (4.5\%).

\subsection{Practical Aspects of MMD Evaluation}
\label{sec:MMD-practical}

An important issue when using the MMD estimator with a Gaussian kernel in practice is the selection of the kernel bandwidth. To mitigate the necessity of tuning $\sigma$ for each experiment, it is usual to rely on a heuristic \cite{gretton2012kernel}. In this study, we rely on the mean heuristic which consists in setting $\sigma$ to be the pairwise mean distance on the aggregate sample, i.e. $\sigma^2 = \frac{\sum_{x, x'} ||x - x'||^2}{N^2 - N}$.

Moreover, instead of a single kernel, we use a mixture of $M$ kernels as in Li et al. (\citeyear{li2015generative}) to make MMD more adaptive and robust to distributional differences:

\begin{equation}
\label{eq:mixture-of-kernels}
k(x, x') = \sum_{i=1}^M k_{\sigma_i}(x, x'),
\end{equation}

\noindent where $k_{\sigma_i}$ represents a Gaussian kernel with bandwidth $\sigma_i$. We set $M = 5$ and $\sigma_i = \alpha_i \sigma$, with $\alpha_i \in \left[\frac{1}{8}, \frac{1}{4}, \frac{1}{2}, 1.0, 2.0 \right]$. 

The computational complexity of calculating the unbiased $\mathrm{MMD}^2$ estimator with a mixture of $M$ Gaussian kernels over a batch of $b$ observations in $\mathbb{R}^d$ is $\mathcal{O}(b^2(d + M))$ when the bandwidth $\sigma$ is estimated using the mean heuristic. The computation of all pairwise squared Euclidean distances dominates the bandwidth calculation step with a cost of $\mathcal{O}(b^2 d)$. The subsequent evaluation of $M$ kernels per pair adds $\mathcal{O}(b^2 M)$. In terms of memory costs, storing $b$ input data in $\mathbb{R}^d$ and $b^2$ pairwise distances yields a complexity of $\mathcal{O}(bd + b^2)$.

Large batch sizes may thus incur significant memory and computational costs during training but improve the accuracy of $\mathrm{MMD}^2$ estimates. Based on available resources, we used large batches ($b \geq 8192$) for \textsc{GloVe} and \textsc{Bias in Bios}, and a moderate size ($b = 2048$) for \textsc{DIAL}. Smaller batches remain feasible, albeit with a slight reduction in performance. The above MMD design choices will be further discussed in Section \ref{sec:res-MMD-practical}.

\subsection{Evaluation metrics}
\label{sec:evaluation-metrics}

\paragraph{Erasure evaluation.} Following prior works \cite{elazar2018adversarial, ravfogel2022linear, basu2024robust}, we evaluate the quality of concept erasure by probing the learned representations for the target concept using nonlinear classifiers. A decrease in probing accuracy is an indicator of erasure. The closer you get to the accuracy of a random predictor, the better the erasure is presumed to be. We also report the Minimum Description Length (MDL) for predicting the erased concept after erasure which has been shown to be a reliable evaluation metric for erasure \cite{voita2020information, basu2022learning}. MDL quantifies the \textit{effort} required by a probing model to achieve a given performance. Higher MDL values suggest more effective erasure.

\paragraph{Quantifying Post-Erasure Utility of the Representations.} When there is no predefined downstream task for which we can evaluate the accuracy, to measure the overall preservation of the semantic information in the embedding space we rely on the metric $A_{k}$ introduced in \cite{basu2024robust} that quantifies the average overlap between the $k$-nearest neighbor sets of $x_i$ and $x_i^\perp$. Formally,  

\begin{equation}
    A_{k} = \frac{1}{N}\sum_{x\in \mathcal{D}} \left[ \frac{1}{k} |k\mathrm{nn}(x) \cap k\mathrm{nn}(x^\perp)| \right]\,,
\end{equation}

\noindent where $k\mathrm{nn}(\cdot)$ computes the $k$-nearest neighbor set of a representation. The metric ranges from $0$ to $1$, with values closer to $1$ indicating greater preservation of the local geometry of the embeddings. We set $k = \frac{|\mathcal{D}|}{2}$, denoting the average neighborhood overlap as $A_{50\%}$.

For datasets of words, like \textsc{GloVe}, WordSim-353 (WS-353) Similarity Evaluation is a standard benchmark for assessing semantic representations by comparing model-predicted word similarities to human-annotated scores on 353 word pairs \cite{agirre2009study}. Model evaluation involves computing cosine similarity between word embeddings and correlating these scores with human ratings using Spearman's rank correlation. For our analysis, we evaluated a subset of 196 word pairs where both words belong to the \textsc{GloVe} vocabulary of the 150k most frequent words and have human similarity ratings available.

\paragraph{Downstream Fairness Evaluation.} A direct application of concept erasure is the improvement of fairness in downstream classifiers by training them on post-erasure representations. To measure downstream classifier bias in \textsc{Bias in Bios} and \textsc{DIAL}, we follow De-Arteaga et al. (\citeyear{dearteaga2019bias}) and report the TPR-Gap metric, which quantifies bias by computing the difference (Gap) in the true positive rate (TPR) between individuals with different groups. Formally, let $Z$,$Y$, and $\hat{Y}$ be random variables denoting respectively a binary protected attribute taking values in $\{c,c'\}$, the downstream task label taking values in $\mathcal{Y}$, and a model's prediction on this task. Then the TPR-Gap $\mathrm{GAP}^{\mathrm{TPR}}_{c;y}$ for a downstream task label $y$ and a protected concept label $c$ is: 

\begin{equation}
\begin{split}
    \mathrm{TPR}_{c,y} = p(\hat{Y} = y | Z = c, Y = y) \\
    \mathrm{GAP}^{\mathrm{TPR}}_{c;y} = \mathrm{TPR}_{c,y} - \mathrm{TPR}_{c',y} 
\end{split}
\end{equation}

In \textsc{Bias in Bios} (resp. \textsc{DIAL}), the concept to erase is the binary gender (resp. race) and $y$ is a profession (resp. a sentiment). Additionally, we consider $\mathrm{GAP}^{\mathrm{TPR, RMS}}_{c}$ as the root mean square (RMS) of the TPR-Gap across all labels in $\mathcal{Y}$ for a given concept class $c$. To explore the relationship between the bias exhibited by the model and the bias in the data, we also compute for \textsc{Bias in Bios} $\sigma_{\mathrm{GAP}^{\mathrm{TPR}}, \%c}$, the correlation between the TPR-Gap for a given profession and the percentage of individual with gender $c$ in that profession.

TPR-GAP may be bad for evaluating fairness in real-world scenarios when $Y$ and $Z$ are highly correlated, as strong debiasing would inevitably lead to a degradation in downstream task performance. For completeness, we also report demographic parity which measures the difference in prediction w.r.t. to a protected attribute: 
\begin{equation}
    \mathrm{DP} = \sum_{y\in \mathcal{Y}} |p(\hat{Y}=y|Z=c) - p(\hat{Y}=y|Z=c')|\\
\end{equation}

\subsection{Implementation Details}
\label{sec:implementation-details}

Key implementation details are summarized below; additional information is provided in Supplementary Material~\ref{app:training-details}. All models are trained on an NVIDIA GeForce RTX 2080 Ti GPU with 11GB of VRAM using CUDA for accelerated computation.

We reimplemented FaRM~\cite{basu2022learning} and KRaM~\cite{basu2024robust}. Concept erasure is performed using a 4-layer MLP with ReLU activations for \textsc{GloVe} and \textsc{Bias in Bios}, and a 7-layer MLP for \textsc{DIAL}. For LEACE, we relied on a publicly available implementation.\footnote{\url{https://github.com/EleutherAI/concept-erasure}}

For \textsc{Bias in Bios}, each input is represented by the final hidden state of a frozen BERT model (bert-base-uncased)~\cite{devlin2019BERT} extracted from the [CLS] token. For \textsc{DIAL}, we use the representations from Basu Roy Chowdhury et al.~(\citeyear{basu2024robust}), obtained from a DeepMoji encoder~\cite{felbo2017using} fine-tuned for sentiment classification.

All models are trained using the Adam optimizer~\cite{kingma2015adam}. For $\overline{\mathrm{L}}$EOPARD, weight decay is set to zero to avoid interfering with the optimization of the projection loss (Eq.~\ref{eq:loss_proj}).

We set $\gamma$ proportionally to $\frac{1}{r^2}$ in the learning objective (Eq.~\ref{eq:total_loss}). In Algorithm \ref{alg:erasure_procedure}, we set $U^{(0)} = \begin{pmatrix} I_r & 0 \end{pmatrix}^\top \in \mathbb{R}^{d \times r}$, a truncated identity matrix. This initialization yields $\mathcal{L}_p(U^{(0)}) = 0$, and we intend to maintain proximity to an orthogonal projection during training.

Evaluation is conducted using scikit-learn’s MLPs~\cite{pedregosa2011scikit} for both probing and downstream classification tasks. All reported accuracies are averaged over five independent runs to ensure robustness.

\section{Results}
\label{sec:results}

\subsection{Nonlinear Concept Erasure}
\label{sec:nonlinear-concept-erasure}

$\overline{\mathrm{L}}$EOPARD enables effective erasure of information associated with a discrete concept, with the projector rank $r$ serving as a tunable parameter that controls the strength of the erasure. Empirically, we observe that the difficulty of recovering the concept post-erasure increases steadily as the number of preserved dimensions decreases (see Figure \ref{fig:tradeoff-mdl-preservation}, left panels). In Table \ref{tab:erasure}, we exhibit configurations where $\overline{\mathrm{L}}$EOPARD outperforms nonlinear erasure baselines in terms of MDL or probing accuracy which drops to near-random levels. Notably, in Table \ref{tab:erasure-glove}, gender treated as a ternary concept during training and evaluation is almost fully erased, demonstrating the validity of $\overline{\mathrm{L}}$EOPARD beyond binary settings. Qualitatively, the visualizations in figure \ref{fig:tsne-glove} and in figures \ref{fig:app_tsne-biasbios},\ref{fig:app_tsne-dial} in Supplementary Material \ref{app:supplementary-results} illustrate that concept class groups, which were previously distinguishable, become indistinguishable following erasure via $\overline{\mathrm{L}}$EOPARD.

\begin{figure}[t!!]
    \centering
    \includegraphics[angle=0,origin=c,width=0.45\columnwidth]{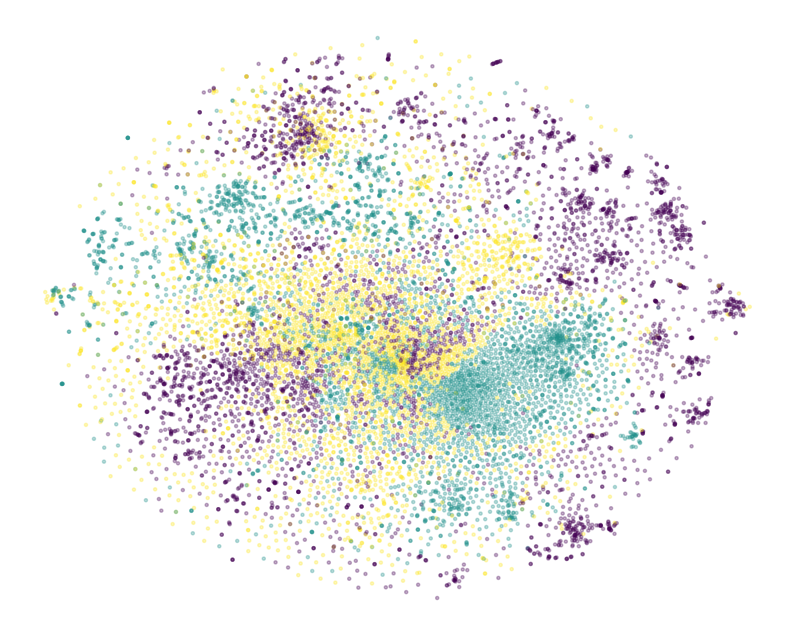}
    \includegraphics[angle=0,origin=c,width=0.45\columnwidth]{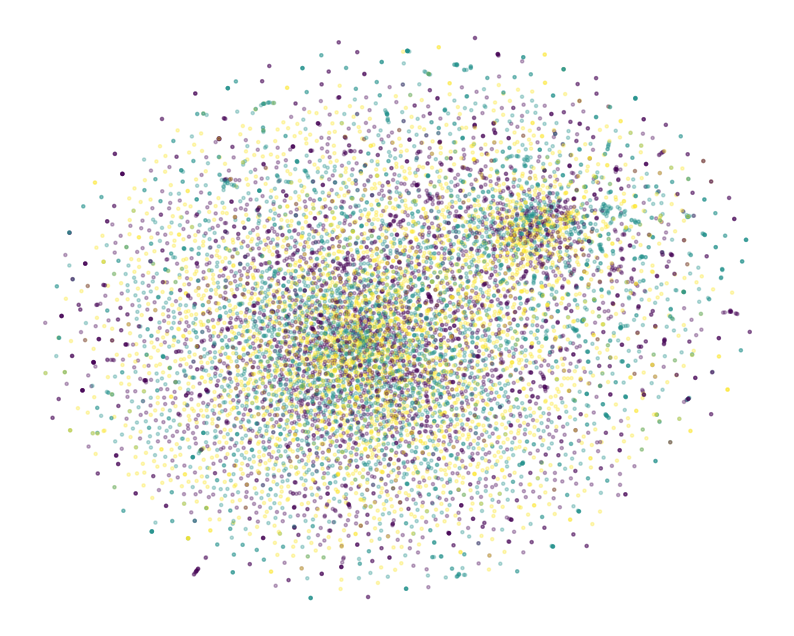}
    \caption{t-SNE visualization of \textsc{GloVe} embeddings before (left) and after (right) erasure via Cascaded $\overline{\mathrm{L}}$EOPARD with a rank-150 orthogonal projection. Gender is a ternary concept. Male-biased, female-biased, and neutral words are displayed in different colors.}
    \label{fig:tsne-glove}
    \par\bigskip
    \par\bigskip
\end{figure}

Results reported in the left panels of Figure \ref{fig:tradeoff-mdl-preservation} highlights the superior erasure capabilities of Cascaded $\overline{\mathrm{L}}$EOPARD (Algorithm \ref{alg:cascaded_erasure_procedure}). For a fixed projection rank, the difficulty of recovering the concept post-erasure is consistently higher with Cascaded $\overline{\mathrm{L}}$EOPARD compared to the standard training (Algorithm \ref{alg:erasure_procedure}). For a given value of MDL, Cascaded $\overline{\mathrm{L}}$EOPARD retains a larger number of dimensions in the feature space. This confirms our intuition that $\overline{\mathrm{L}}$EOPARD trained in a standard way struggles with the removal of strong linear signals, which undermines nonlinear erasure performance. However, concept erasure is achieved with standard $\overline{\mathrm{L}}$EOPARD at low rank. Both FaRM and KRaM, in their standard forms, perform well in terms of both linear and nonlinear erasure, suggesting that implementing them in a cascaded setup should not improve their performance, which we confirm experimentally (see Supplementary Material \ref{app:supplementary-results}).

\begin{table}[t!]
\caption{Concept erasure evaluation. Accuracies are reported in $\%$ and MDL in kBits.}
\label{tab:erasure}
\par\bigskip 
\centering
\begin{subtable}[t]{\columnwidth}
\caption{Gender erasure in \textsc{GloVe} embeddings. \scriptsize{$a_z^{\mathrm{bin}}$ (resp. $a_z^{\mathrm{ter}}$) denotes the accuracy to predict the gender considered as binary (resp. ternary)}.}
\label{tab:erasure-glove}
\centering
\resizebox{\linewidth}{!}{%
\begin{tabular}{llclcc}
\toprule
 & $a_z^{\mathrm{bin}}$ $\downarrow$ & MDL $\uparrow$ \scriptsize & $a_z^{\mathrm{ter}}$ $\downarrow$ & $A_{50\%} \uparrow$ & WS-353 $\uparrow$\\ 
\midrule
\textit{original}   & \textit{100.0\;\scriptsize$\pm$0.0} & \textit{0.1} &  \textit{100.0\;\scriptsize$\pm$0.0} & \textit{1.00} & \textit{0.70} \\
\textit{random}     & \textit{50.0} & -- & \textit{33.3} & \textit{0.50} & --\vspace{0.5em}\\

FaRM & \textbf{52.7\;\scriptsize$\pm$0.4} & 25.3 & \textbf{36.4\;\scriptsize$\pm$0.6} &  0.67 & 0.57\\ 
KRaM & 53.7\;\scriptsize$\pm$0.4 & 27.9 & \textbf{36.1\;\scriptsize$\pm$0.5} &  0.62 & 0.58\vspace{0.5em}\\ 
\rowcolor{LightLightGray}$\overline{\mathrm{L}}$EOPARD& & & & &  \\
\rowcolor{LightLightGray}\scriptsize standard, r=75 & 57.3\;\scriptsize$\pm$0.6 & 26.2 & 39.7\;\scriptsize$\pm$0.4 & 0.75  & 0.45\\
\rowcolor{LightLightGray}\scriptsize cascaded, r=125 & \textbf{51.8\;\scriptsize$\pm$0.9} & \textbf{30.4} & \textbf{36.3\;\scriptsize$\pm$0.5} & \textbf{0.76}  & \textbf{0.64}  \\

\bottomrule
\end{tabular}
}
\end{subtable}
\par\bigskip 

\begin{subtable}[t]{\columnwidth}
\caption{Gender erasure in \textsc{Bias in Bios}.}
\label{tab:erasure-biasbios}
\centering
\resizebox{0.7\columnwidth}{!}{%

\begin{tabular}{llll}
\toprule
 & $a_z$ $\downarrow$ & MDL $\uparrow$  & $a_y$ $\uparrow$\\ 
\midrule
\textit{original}   & \textit{99.4\;\scriptsize $\pm$0.0} & 2.7 & \textit{80.0\;\scriptsize$\pm$\;0.1}   \\
\textit{random}     & \textit{53.5} & -- & \textit{33.5}\vspace{0.5em}\\ 

FaRM & 58.8\;\scriptsize$\pm$\;0.2 & \textbf{236.9} &  55.6\;\scriptsize$\pm$\;0.1\\
KRaM & 56.4\;\scriptsize$\pm$\;0.3 & 211.8 &  51.4\;\scriptsize$\pm$\;0.1 \vspace{0.5em}\\

\rowcolor{LightLightGray}$\overline{\mathrm{L}}$EOPARD & & & \\  

\rowcolor{LightLightGray}\scriptsize standard, r=10 & 58.8\;\scriptsize$\pm$\;0.7 & 193.4 & 53.4\;\scriptsize$\pm$\;0.1 \\
\rowcolor{LightLightGray}\scriptsize cascaded, r=50 & \textbf{54.1\;\scriptsize$\pm$\;0.2} & 193.5 & 54.1\;\scriptsize$\pm$\;0.6 \\
\rowcolor{LightLightGray}\scriptsize cascaded, r=150 & 55.9\;\scriptsize$\pm$\;0.3 & 192.2 & \textbf{65.5\;\scriptsize$\pm$\;0.5} \\
\bottomrule
\end{tabular}
}
\end{subtable}
\par\bigskip

\begin{subtable}[t]{\columnwidth}
\caption{Race erasure in \textsc{DIAL}.}
\label{tab:erasure-dial}
\centering
\resizebox{0.7\columnwidth}{!}{%
\begin{tabular}{llll}
\toprule
 & $a_z$ $\downarrow$ & MDL $\uparrow$ &  $a_y$ $\uparrow$\\ 
\midrule
\textit{original}   & \textit{88.0\;\scriptsize$\pm$\;0.1} & 136.2 & \textit{75.8\;\scriptsize$\pm$\;0.1}\\
\textit{random}     & \textit{50.0} & -- & \textit{33.3}\vspace{0.5em}\\ 

FaRM & \textbf{53.7\;\scriptsize$\pm$\;0.8} & \textbf{335.8} & \textbf{73.9\;\scriptsize$\pm$\;0.3}\\
KRaM & \textbf{53.7\;\scriptsize$\pm$\;0.5} & 330.3 & \textbf{73.9\;\scriptsize$\pm$\;0.2}\vspace{0.5em}\\ 

\rowcolor{LightLightGray}$\overline{\mathrm{L}}$EOPARD & & & \\  

\rowcolor{LightLightGray}\scriptsize standard, r=15 & 56.8\;\scriptsize$\pm$\;0.2 & 311.3 &   67.4\;\scriptsize$\pm$\;0.1\\
\rowcolor{LightLightGray}\scriptsize cascaded, r=15 & 57.2\;\scriptsize$\pm$\;0.3 & 309.0 &   70.2\;\scriptsize$\pm$\;0.1\\

\bottomrule
\end{tabular}
}
\end{subtable}

\end{table}

\subsection{Erasure vs. utility tradeoff}
\label{sec:erasure-utility-tradeoff}

\begin{figure}[t!]
    \centering
     \begin{subfigure}[b]{\linewidth}
         \centering
         \includegraphics[width=0.8\linewidth]{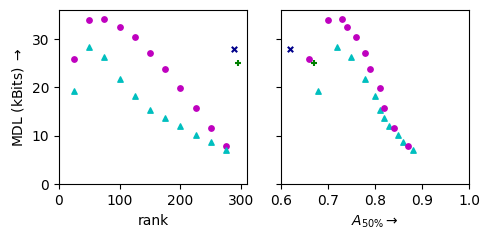}
         \vspace{-1\baselineskip}
         \caption{\textsc{GloVe}}
         \label{fig:tradeoff-glove}
     \end{subfigure}
     \par\bigskip 
     \begin{subfigure}[b]{\linewidth}
         \centering
         \includegraphics[width=0.8\linewidth]{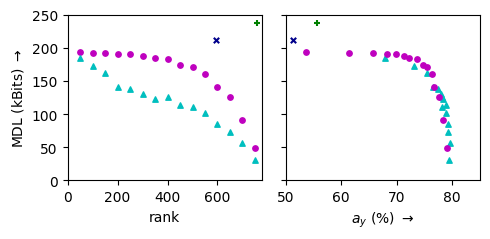}
         \vspace{-1\baselineskip}
         \caption{\textsc{Bias in Bios}}
         \label{fig:tradeoff-biasbios}
     \end{subfigure}
     \par\bigskip 
     \begin{subfigure}[b]{\linewidth}
         \centering
         \includegraphics[width=0.8\linewidth]{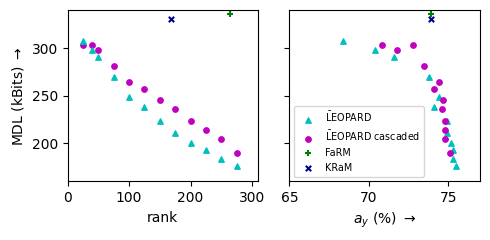}
         \vspace{-1\baselineskip}
         \caption{\textsc{DIAL}}
         \label{fig:tradeoff-dial}
     \end{subfigure}
     \par\bigskip 
    \caption{Erasure vs. rank of the feature space post-erasure (left) and Erasure–Utility trade-off (right) for each dataset. The vertical axis, erasure strength as evaluated by MDL, is shared between the right and left panels.}
    \label{fig:tradeoff-mdl-preservation}
    \par\bigskip
    \par\bigskip
\end{figure}

The right panels of Figure~\ref{fig:tradeoff-mdl-preservation} highlight a consistent trade-off between concept erasure and the utility of post-erasure representations across all benchmark datasets, with optimal configurations located in the upper-right regions. The rank $r$, which controls the strength of erasure, modulates this trade-off. FaRM and KRaM achieve strong erasure but at the cost of substantial degradation in utility for gender, as evidenced by their positions in the upper-left regions of the left panels in Figures~\ref{fig:tradeoff-glove} and \ref{fig:tradeoff-biasbios}. 

The adjustable rank of $\overline{\mathrm{L}}$EOPARD offers flexibility in navigating this trade-off, enabling configurations that outperform baselines across several settings. Notably, on \textsc{GloVe} embeddings, $\overline{\mathrm{L}}$EOPARD better preserves both local neighborhood structure and semantic content (Table~\ref{tab:erasure-glove}).

On \textsc{Bias in Bios}, FaRM and KRaM outperform $\overline{\mathrm{L}}$EOPARD in terms of MDL-based erasure at the expense of downstream task performance. At low rank, $\overline{\mathrm{L}}$EOPARD achieves trade-offs between probing accuracy and downstream task accuracy comparable to baselines (e.g., $r=10$ and $r=50$ for standard and cascaded training, respectively, in Table~\ref{tab:erasure-biasbios}). Figure~\ref{fig:tradeoff-biasbios} reveals that beyond a certain threshold, performance plateaus; further reducing $r$ yields marginal gains in erasure while significantly harming utility. In such cases, $\overline{\mathrm{L}}$EOPARD (e.g., at $r=150$ with cascaded training) allows for a more satisfying balance between the reduction of the probing accuracy and downstream accuracy compared to baselines.

On \textsc{DIAL}, for race erasure, $\overline{\mathrm{L}}$EOPARD remains competitive without outperforming the baselines. Relying on linear projections in $\overline{\mathrm{L}}$EOPARD provides advantages in simplicity, interpretability, and parameter efficiency. However, such linear methods are inherently limited when addressing the erasure of concepts encoded nonlinearly. While erasing even a small number of dimensions can significantly reduce the recoverability of the target concept, achieving strong erasure often requires neutralizing a large subspace, which may degrade the utility of the resulting representations.

\subsection{Fairness Analysis}
\label{sec:fairness-analysis}

Fairness evaluation results, reported in Table~\ref{tab:fairness-analysis}, demonstrate that $\overline{\mathrm{L}}$EOPARD significantly reduces gender and racial bias in downstream classification tasks, achieving performance comparable to the strongest baselines at low rank (e.g., $r = 10$ and $r=50$ with standard and cascaded training, respectively). The bias metrics consistently decrease as the number of preserved dimensions in the feature space is reduced, indicating an increasing level of bias mitigation.

\begin{figure}[t!]
    \centering
    \includegraphics[width=0.85\linewidth]{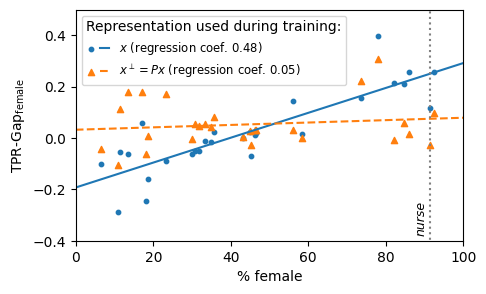}
    \caption{Average $\mathrm{TPR}\text{-Gap}_{\text{female}, y}$ vs. female proportion in each occupation $y$ for different representations used train the classifiers. $P$ is the rank-250 projection calculated via Cascaded $\overline{\mathrm{L}}$EOPARD. Each vertical corresponds to a specific occupation (e.g., nurse).}
    \label{fig:biasbios-fairness}
    \par\bigskip
    \par\bigskip
\end{figure}

Table~\ref{tab:fairness-analysis-biasbios} highlights a trade-off specific to the \textsc{Bias in Bios} dataset between classifier performance on the downstream task (occupation prediction) and classifier fairness.\footnote{Such a trade-off has been demonstrated by Zaho \& Gordon (\citeyear{zhao2022inherent})} For FaRM and KRaM, achieving fairer classifiers comes at the cost of a substantial drop in downstream task accuracy. In contrast, the tunable rank in $\overline{\mathrm{L}}$EOPARD enables a more flexible compromise—for example, achieving significant gender bias reduction while incurring a smaller loss in downstream performance at $r=350$ or $r=250$ with Cascaded $\overline{\mathrm{L}}$EOPARD. The significant reduction of the positive correlation between the TPR gender gap for an occupation and the gender imbalance in that occupation at $r=250$ is made clearly visible in Figure \ref{fig:biasbios-fairness}.

The results in Table~\ref{tab:fairness-analysis-dial} reveal a more nuanced picture: although $\overline{\mathrm{L}}$EOPARD outperforms the baselines in bias reduction at low rank (e.g. $r=15$), it does so at the expense of lower downstream accuracy. Depending on the application context, particularly when fairness is a primary concern, the ability to adjust the rank makes $\overline{\mathrm{L}}$EOPARD a favorable choice over less flexible methods.

\begin{table}[t!]
\caption{Fairness evaluation of nonlinear downstream classifiers}
\label{tab:fairness-analysis}
\par\bigskip 
\centering
\begin{subtable}[t]{\columnwidth}
\centering
\caption{Occupation prediction on \textsc{Bias in Bios}}
\label{tab:fairness-analysis-biasbios}
\resizebox{\columnwidth}{!}{%
\begin{tabular}{lllll}
\toprule
 & $a_y$ $\uparrow$ \scriptsize(\%) & $\mathrm{GAP}^{\mathrm{TPR, RMS}}_{\mathrm{female}}$ $\downarrow$ &  $\sigma_{\mathrm{GAP}, \%\mathrm{female}}$ $\downarrow$ & DP$\downarrow$ \\
\midrule
\textit{original}   & \textit{79.5\;\scriptsize$\pm$\;0.2}& \textit{0.156\;\scriptsize$\pm$\;0.006} &  \textit{0.84\;\scriptsize$\pm$\;0.029} &  \textit{0.57\;\scriptsize$\pm$\;0.009}\vspace{0.5em}\\ 

FaRM & 55.6\;\scriptsize$\pm$\;0.1 & 0.073\;\scriptsize$\pm$\;0.004 & 0.31\;\scriptsize$\pm$\;0.021  & 0.33\;\scriptsize$\pm$\;0.004\\ 
KRaM & 51.4\;\scriptsize$\pm$\;0.1 & \textbf{0.065\;\scriptsize$\pm$\;0.003} &  \textbf{0.085\;\scriptsize$\pm$\;0.087} & \textbf{0.28\;\scriptsize$\pm$\;0.005}\vspace{0.5em}\\ 
\rowcolor{LightLightGray}$\overline{\mathrm{L}}$EOPARD &  & & & \\

\rowcolor{LightLightGray}\scriptsize cascaded, r=350 &  \textbf{72.7\;\scriptsize$\pm$\;0.3} & 0.104\;\scriptsize$\pm$\;0.007 & 0.27\;\scriptsize$\pm$\;0.033 & 0.43\;\scriptsize$\pm$\;0.002\\ 
\rowcolor{LightLightGray}\scriptsize cascaded, r=250 & 70.1\;\scriptsize$\pm$\;0.4 & 0.106\;\scriptsize$\pm$\;0.004 & 0.13\;\scriptsize$\pm$\;0.050 & 0.40\;\scriptsize$\pm$\;0.006\\ 
\rowcolor{LightLightGray}\scriptsize cascaded, r=50 & 54.1\;\scriptsize$\pm$\;0.6 & 0.082\;\scriptsize$\pm$\;0.004 & \textbf{0.0079\;\scriptsize$\pm$\;0.075} & \textbf{0.29\;\scriptsize$\pm$\;0.007}\\ 
\rowcolor{LightLightGray}\scriptsize standard, r=10 & 53.4\;\scriptsize$\pm$\;0.1 & \textbf{0.066\;\scriptsize$\pm$\;0.001} & \textbf{0.097\;\scriptsize$\pm$\;0.052}  & 0.29\;\scriptsize$\pm$\;0.002 \\ 
\bottomrule
\end{tabular}
}
\end{subtable}
\par\bigskip 

\begin{subtable}[t]{\columnwidth}
\centering
\caption{Sentiment prediction on \textsc{DIAL}}
\label{tab:fairness-analysis-dial}
\resizebox{0.8\columnwidth}{!}{%
\begin{tabular}{lllll}

\toprule
 & $a_y$ $\uparrow$ \scriptsize(\%) & $\mathrm{GAP}^{\mathrm{TPR, RMS}}_{\%\mathrm{AAE}}$ $\downarrow$ &  DP$\downarrow$ \\
\midrule
\textit{original}   & \textit{75.8\;\scriptsize$\pm$\;0.1}& \textit{0.150\;\scriptsize$\pm$\;0.003} &   \textit{0.258\;\scriptsize$\pm$\;0.007}\vspace{0.5em}\\ 

FaRM & \textbf{73.9\;\scriptsize$\pm$\;0.3} & 0.071\;\scriptsize$\pm$\;0.005  & 0.060\;\scriptsize$\pm$\;0.005\\ 
KRaM & \textbf{73.9\;\scriptsize$\pm$\;0.2} & 0.066\;\scriptsize$\pm$\;0.003 &  0.039\;\scriptsize$\pm$\;0.010\vspace{0.5em}\\ 
\rowcolor{LightLightGray}$\overline{\mathrm{L}}$EOPARD &  & & \\
\rowcolor{LightLightGray}\scriptsize cascaded, r=75 & \textbf{73.5\;\scriptsize$\pm$\;0.2} & 0.101\;\scriptsize$\pm$\;0.002 &  0.122\;\scriptsize$\pm$\;0.009\\
\rowcolor{LightLightGray}\scriptsize cascaded, r=50 &  71.8\;\scriptsize$\pm$\;0.2 & 0.069\;\scriptsize$\pm$\;0.004 &  0.071\;\scriptsize$\pm$\;0.005\\ 
\rowcolor{LightLightGray}\scriptsize cascaded, r=15 &  70.2\;\scriptsize$\pm$\;0.1 & \textbf{0.051\;\scriptsize$\pm$\;0.005} &  0.069\;\scriptsize$\pm$\;0.009\\ 
\rowcolor{LightLightGray}\scriptsize standard, r=15 &  67.4\;\scriptsize$\pm$\;0.1 & \textbf{0.042\;\scriptsize$\pm$\;0.004} &  \textbf{0.022\;\scriptsize$\pm$\;0.003} \\
\bottomrule
\end{tabular}
}
\end{subtable}
\end{table}

\subsection{About MMD design choices}
\label{sec:res-MMD-practical}

This work proposes an implementation of $\overline{\mathrm{L}}$EOPARD designed for broad applicability with minimal hyperparameter tuning. As detailed in Section~\ref{sec:MMD-practical}, several choices regarding the evaluation of MMD were informed by prior studies and supported by empirical evidence. Specifically, employing a single kernel ($M=1$) yields results of comparable magnitude to those reported in this article, while increasing $M$ beyond 2 kernels yields only marginal improvements (see Supplementary Material~\ref{app:kernel-analysis}). We set $M=5$ in all experiments as a trade-off between performance and computational cost. 

Regarding batch size, empirical analyses in Supplementary Material~\ref{app:batch-size-analysis} indicate that large batches ($b \geq 8192$) are preferable for gender erasure, while moderate batch sizes ($b = 2048$) are well-suited for race erasure. Nevertheless, smaller batches ($b = 512$) remain effective, often achieving competitive trade-offs and, in certain configurations, surpassing state-of-the-art baselines.

\subsection{About the freedom of choice for the rank}
\label{sec:choice-of-r}

There is no universally optimal rank $r$ for concept erasure with $\overline{\mathrm{L}}$EOPARD. The ability to adjust the rank is a key strength of the method, allowing for context-specific trade-offs between erasure effectiveness and post-erasure representation utility. In practice, several strategies can guide the selection of an appropriate rank.

A straightforward approach involves defining a threshold on an erasure metric (e.g., MDL or validation accuracy), or more generally, on any task-independent criterion in unconstrained settings, and selecting the largest rank that satisfies this constraint. Alternatively, one may plot erasure and utility metrics as a function of $r$, similar to the analysis presented in Figure~\ref{fig:tradeoff-mdl-preservation}, to identify plateaus or local optima, providing a graphical heuristic for selecting a suitable rank.

\section{Scope and Limitations}
\label{sec:limitations}

Our method specifically targets the erasure of discrete concepts. It could be naturally extended to the erasure of multiple discrete concepts by replacing the \emph{erasure loss} (Eq.~\ref{eq:mmd_loss_multi}) with a sum of individual erasure losses, one per concept to be erased, in Eq.~\ref{eq:total_loss}. With $\overline{\mathrm{L}}$EOPARD, we do not address the erasure of continuous concepts, as the density matching objective in Eq.~\ref{eq:mmd_loss_multi} requires partitioning samples according to discrete concept classes. For continuous concept erasure, approaches such as FaRM \cite{basu2022learning} or KRaM \cite{basu2024robust} are more suitable.

While our approach currently relies on orthogonal projections to ensure the preservation of local geometric structure and to avoid trivial solutions (e.g., constant mappings), this restriction limits the expressiveness of the method. Experiments on the \textsc{DIAL} dataset highlight this limitation: effective neutralization requires removing a large number of dimensions, which adversely impacts information preservation. To address this, future work will investigate integrating nonlinear erasure functions with the density matching objective.

\section{Conclusion and future work}
\label{sec:conclusion}
We introduced $\overline{\mathrm{L}}$EOPARD, a simple yet effective method for nonlinear concept erasure which uses orthogonal projections to neutralize concept-specific information. By optimizing a density matching objective, $\overline{\mathrm{L}}$EOPARD ensures indistinguishability of class-conditional distributions in the the embedding space after erasure. Evaluated on classic NLP benchmarks, $\overline{\mathrm{L}}$EOPARD is competitive and even outperforms state-of-the-art methods in nonlinear erasure of demographic attributes while better preserving the post-erasure utility of the representations. A notable strength of $\overline{\mathrm{L}}$EOPARD is its ability to control the trade-off between concept erasure and information preservation by adjusting the rank of the learned projection. Empirical results also demonstrate that $\overline{\mathrm{L}}$EOPARD improves fairness in downstream classifiers with a moderate impact on task performance.

Future research could explore applying the class-conditional density matching objective within an end-to-end model training alongside a generic objective such as MLM for BERT fine-tuning. Additionally, our approach could provide new insights into fairness and causal effect evaluation.

\clearpage

% \section*{Impact Statement and Ethical Considerations}

% This work explores the removal of demographic attributes information from pre-trained representations, with potential applications in fairness. However, we caution that the method should not be seen as a definitive solution to bias in neural models. Its effectiveness depends on context, including the data and fairness metrics in use. The experiments are limited to a specific set of datasets, which may not capture all forms of implicit gender bias, meaning some biases may persist post-application. Additionally, the method is tailored to a particular technical definition of bias and may not be robust to other forms of bias or adversarial approaches. We encourage future research to address these limitations and further investigate its applicability.

%%%%%%%%%%%%%%%%%%%%%%%%%%%%%%%%%%%%%%%%%%%%%%%%%%%%%%%%%%%%%%%%%%%%%%%%

%%% Use this command to include your bibliography file.

\bibliography{bibliography}

\begin{thebibliography}{30}
\providecommand{\natexlab}[1]{#1}
\providecommand{\url}[1]{\texttt{#1}}
\expandafter\ifx\csname urlstyle\endcsname\relax
  \providecommand{\doi}[1]{doi: #1}\else
  \providecommand{\doi}{doi: \begingroup \urlstyle{rm}\Url}\fi

\bibitem[Abraham et~al.(2022)Abraham, D'Oosterlinck, Feder, Gat, Geiger, Potts, Reichart, and Wu]{abraham2022cebab}
E.~D. Abraham, K.~D'Oosterlinck, A.~Feder, Y.~Gat, A.~Geiger, C.~Potts, R.~Reichart, and Z.~Wu.
\newblock {CEBaB}: Estimating the causal effects of real-world concepts on nlp model behavior.
\newblock \emph{Advances in Neural Information Processing Systems}, 35:\penalty0 17582--17596, 2022.

\bibitem[Agirre et~al.(2009)Agirre, Alfonseca, Hall, Kravalova, Pa{\c{s}}ca, and Soroa]{agirre2009study}
E.~Agirre, E.~Alfonseca, K.~Hall, J.~Kravalova, M.~Pa{\c{s}}ca, and A.~Soroa.
\newblock A study on similarity and relatedness using distributional and {W}ord{N}et-based approaches.
\newblock In M.~Ostendorf, M.~Collins, S.~Narayanan, D.~W. Oard, and L.~Vanderwende, editors, \emph{Proceedings of Human Language Technologies: The 2009 Annual Conference of the North {A}merican Chapter of the Association for Computational Linguistics}, pages 19--27, Boulder, Colorado, June 2009. Association for Computational Linguistics.

\bibitem[Basu Roy~Chowdhury and Chaturvedi(2022)]{basu2022learning}
S.~Basu Roy~Chowdhury and S.~Chaturvedi.
\newblock Learning fair representations via rate-distortion maximization.
\newblock \emph{Transactions of the Association for Computational Linguistics}, 10:\penalty0 1159--1174, 2022.
\newblock \doi{10.1162/tacl_a_00512}.

\bibitem[Basu Roy~Chowdhury et~al.(2023)Basu Roy~Chowdhury, Monath, Dubey, Ahmed, and Chaturvedi]{basu2024robust}
S.~Basu Roy~Chowdhury, N.~Monath, K.~A. Dubey, A.~Ahmed, and S.~Chaturvedi.
\newblock Robust concept erasure via kernelized rate-distortion maximization.
\newblock \emph{Advances in Neural Information Processing Systems}, 36, 2023.

\bibitem[Belrose et~al.(2023)Belrose, Schneider-Joseph, Ravfogel, Cotterell, Raff, and Biderman]{belrose2023leace}
N.~Belrose, D.~Schneider-Joseph, S.~Ravfogel, R.~Cotterell, E.~Raff, and S.~Biderman.
\newblock {LEACE}: Perfect linear concept erasure in closed form.
\newblock In \emph{Thirty-seventh Conference on Neural Information Processing Systems}, 2023.

\bibitem[Blodgett et~al.(2016)Blodgett, Green, and O’Connor]{blodgett2016demographic}
S.~L. Blodgett, L.~Green, and B.~O’Connor.
\newblock Demographic dialectal variation in social media: A case study of african-american english.
\newblock In \emph{Proceedings of the 2016 Conference on Empirical Methods in Natural Language Processing}, pages 1119--1130, 2016.

\bibitem[Bolukbasi et~al.(2016)Bolukbasi, Chang, Zou, Saligrama, and Kalai]{bolukbasi2016man}
T.~Bolukbasi, K.-W. Chang, J.~Y. Zou, V.~Saligrama, and A.~T. Kalai.
\newblock Man is to computer programmer as woman is to homemaker? debiasing word embeddings.
\newblock \emph{Advances in neural information processing systems}, 29, 2016.

\bibitem[Brown et~al.(2020)Brown, Mann, Ryder, Subbiah, Kaplan, Dhariwal, Neelakantan, Shyam, Sastry, Askell, Agarwal, Herbert-Voss, Krueger, Henighan, Child, Ramesh, Ziegler, Wu, Winter, Hesse, Chen, Sigler, Litwin, Gray, Chess, Clark, Berner, McCandlish, Radford, Sutskever, and Amodei]{brown2020language}
T.~Brown, B.~Mann, N.~Ryder, M.~Subbiah, J.~D. Kaplan, P.~Dhariwal, A.~Neelakantan, P.~Shyam, G.~Sastry, A.~Askell, S.~Agarwal, A.~Herbert-Voss, G.~Krueger, T.~Henighan, R.~Child, A.~Ramesh, D.~Ziegler, J.~Wu, C.~Winter, C.~Hesse, M.~Chen, E.~Sigler, M.~Litwin, S.~Gray, B.~Chess, J.~Clark, C.~Berner, S.~McCandlish, A.~Radford, I.~Sutskever, and D.~Amodei.
\newblock Language models are few-shot learners.
\newblock In H.~Larochelle, M.~Ranzato, R.~Hadsell, M.~Balcan, and H.~Lin, editors, \emph{Advances in Neural Information Processing Systems}, volume~33, pages 1877--1901. Curran Associates, Inc., 2020.

\bibitem[Caliskan et~al.(2017)Caliskan, Bryson, and Narayanan]{caliskan2017semantics}
A.~Caliskan, J.~J. Bryson, and A.~Narayanan.
\newblock Semantics derived automatically from language corpora contain human-like biases.
\newblock \emph{Science}, 356\penalty0 (6334):\penalty0 183--186, 2017.
\newblock \doi{10.1126/science.aal4230}.

\bibitem[De-Arteaga et~al.(2019)De-Arteaga, Romanov, Wallach, Chayes, Borgs, Chouldechova, Geyik, Kenthapadi, and Kalai]{dearteaga2019bias}
M.~De-Arteaga, A.~Romanov, H.~Wallach, J.~Chayes, C.~Borgs, A.~Chouldechova, S.~Geyik, K.~Kenthapadi, and A.~T. Kalai.
\newblock Bias in bios: A case study of semantic representation bias in a high-stakes setting.
\newblock In \emph{Proceedings of the Conference on Fairness, Accountability, and Transparency}, FAT* '19, page 120–128, New York, NY, USA, 2019. Association for Computing Machinery.
\newblock ISBN 9781450361255.
\newblock \doi{10.1145/3287560.3287572}.

\bibitem[Devlin et~al.(2019)Devlin, Chang, Lee, and Toutanova]{devlin2019BERT}
J.~Devlin, M.-W. Chang, K.~Lee, and K.~Toutanova.
\newblock Bert: Pre-training of deep bidirectional transformers for language understanding.
\newblock In \emph{Proceedings of the 2019 Conference of the North American Chapter of the Association for Computational Linguistics: Human Language Technologies, NAACL-HLT 2019}, volume~1, pages 4171–--4186, Minneapolis, MN, USA, 2019. Association for Computational Linguistics.

\bibitem[Elazar and Goldberg(2018)]{elazar2018adversarial}
Y.~Elazar and Y.~Goldberg.
\newblock Adversarial removal of demographic attributes from text data.
\newblock In E.~Riloff, D.~Chiang, J.~Hockenmaier, and J.~Tsujii, editors, \emph{Proceedings of the 2018 Conference on Empirical Methods in Natural Language Processing}, pages 11--21, Brussels, Belgium, Oct.-Nov. 2018. Association for Computational Linguistics.
\newblock \doi{10.18653/v1/D18-1002}.

\bibitem[Feder et~al.(2021)Feder, Oved, Shalit, and Reichart]{feder2021causalm}
A.~Feder, N.~Oved, U.~Shalit, and R.~Reichart.
\newblock {CausaLM}: Causal model explanation through counterfactual language models.
\newblock \emph{Computational Linguistics}, 47\penalty0 (2):\penalty0 333--386, 2021.

\bibitem[Felbo et~al.(2017)Felbo, Mislove, S{\o}gaard, Rahwan, and Lehmann]{felbo2017using}
B.~Felbo, A.~Mislove, A.~S{\o}gaard, I.~Rahwan, and S.~Lehmann.
\newblock Using millions of emoji occurrences to learn any-domain representations for detecting sentiment, emotion and sarcasm.
\newblock In \emph{Proceedings of the 2017 Conference on Empirical Methods in Natural Language Processing}, pages 1615--1625, 2017.

\bibitem[Ganin et~al.(2016)Ganin, Ustinova, Ajakan, Germain, Larochelle, Laviolette, March, and Lempitsky]{ganin2016domain}
Y.~Ganin, E.~Ustinova, H.~Ajakan, P.~Germain, H.~Larochelle, F.~Laviolette, M.~March, and V.~Lempitsky.
\newblock Domain-adversarial training of neural networks.
\newblock \emph{Journal of machine learning research}, 17\penalty0 (59):\penalty0 1--35, 2016.

\bibitem[Gretton et~al.(2012)Gretton, Borgwardt, Rasch, Sch{{\"o}}lkopf, and Smola]{gretton2012kernel}
A.~Gretton, K.~M. Borgwardt, M.~J. Rasch, B.~Sch{{\"o}}lkopf, and A.~Smola.
\newblock A kernel two-sample test.
\newblock \emph{Journal of Machine Learning Research}, 13\penalty0 (25):\penalty0 723--773, 2012.

\bibitem[Kingma and Ba(2015)]{kingma2015adam}
D.~P. Kingma and J.~Ba.
\newblock Adam: {A} method for stochastic optimization.
\newblock In Y.~Bengio and Y.~LeCun, editors, \emph{3rd International Conference on Learning Representations, {ICLR} 2015, San Diego, CA, USA, May 7-9, 2015, Conference Track Proceedings}, 2015.

\bibitem[Lemberger and Saillenfest(2024)]{lemberger2024explaining}
P.~Lemberger and A.~Saillenfest.
\newblock Explaining text classifiers with counterfactual representations.
\newblock In U.~Endriss, F.~S. Melo, K.~Bach, A.~J.~B. Diz, J.~M. Alonso{-}Moral, S.~Barro, and F.~Heintz, editors, \emph{{ECAI} 2024 - 27th European Conference on Artificial Intelligence, 19-24 October 2024, Santiago de Compostela, Spain - Including 13th Conference on Prestigious Applications of Intelligent Systems {(PAIS} 2024)}, volume 392 of \emph{Frontiers in Artificial Intelligence and Applications}, pages 890--897. {IOS} Press, 2024.
\newblock \doi{10.3233/FAIA240576}.

\bibitem[Li et~al.(2015)Li, Swersky, and Zemel]{li2015generative}
Y.~Li, K.~Swersky, and R.~Zemel.
\newblock Generative moment matching networks.
\newblock In F.~Bach and D.~Blei, editors, \emph{Proceedings of the 32nd International Conference on Machine Learning}, volume~37 of \emph{Proceedings of Machine Learning Research}, pages 1718--1727, Lille, France, 07--09 Jul 2015. PMLR.

\bibitem[Makar et~al.(2022)Makar, Packer, Moldovan, Blalock, Halpern, and D’Amour]{makar2022causally}
M.~Makar, B.~Packer, D.~Moldovan, D.~Blalock, Y.~Halpern, and A.~D’Amour.
\newblock Causally motivated shortcut removal using auxiliary labels.
\newblock In \emph{International Conference on Artificial Intelligence and Statistics}, pages 739--766. PMLR, 2022.

\bibitem[Pedregosa et~al.(2011)Pedregosa, Varoquaux, Gramfort, Michel, Thirion, Grisel, Blondel, Prettenhofer, Weiss, Dubourg, et~al.]{pedregosa2011scikit}
F.~Pedregosa, G.~Varoquaux, A.~Gramfort, V.~Michel, B.~Thirion, O.~Grisel, M.~Blondel, P.~Prettenhofer, R.~Weiss, V.~Dubourg, et~al.
\newblock Scikit-learn: Machine learning in python.
\newblock \emph{the Journal of machine Learning research}, 12:\penalty0 2825--2830, 2011.

\bibitem[Pennington et~al.(2014)Pennington, Socher, and Manning]{pennington2014glove}
J.~Pennington, R.~Socher, and C.~D. Manning.
\newblock Glove: Global vectors for word representation.
\newblock In \emph{Proceedings of the 2014 conference on empirical methods in natural language processing (EMNLP)}, pages 1532--1543, 2014.

\bibitem[Ravfogel et~al.(2020)Ravfogel, Elazar, Gonen, Twiton, and Goldberg]{ravfogel-etal-2020-null}
S.~Ravfogel, Y.~Elazar, H.~Gonen, M.~Twiton, and Y.~Goldberg.
\newblock Null it out: Guarding protected attributes by iterative nullspace projection.
\newblock In D.~Jurafsky, J.~Chai, N.~Schluter, and J.~Tetreault, editors, \emph{Proceedings of the 58th Annual Meeting of the Association for Computational Linguistics}, pages 7237--7256, Online, July 2020. Association for Computational Linguistics.
\newblock \doi{10.18653/v1/2020.acl-main.647}.

\bibitem[Ravfogel et~al.(2022{\natexlab{a}})Ravfogel, Twiton, Goldberg, and Cotterell]{ravfogel2022linear}
S.~Ravfogel, M.~Twiton, Y.~Goldberg, and R.~D. Cotterell.
\newblock Linear adversarial concept erasure.
\newblock In \emph{International Conference on Machine Learning}, pages 18400--18421. PMLR, 2022{\natexlab{a}}.

\bibitem[Ravfogel et~al.(2022{\natexlab{b}})Ravfogel, Vargas, Goldberg, and Cotterell]{ravfogel2022adversarial}
S.~Ravfogel, F.~Vargas, Y.~Goldberg, and R.~Cotterell.
\newblock Adversarial concept erasure in kernel space.
\newblock In \emph{Proceedings of the 2022 Conference on Empirical Methods in Natural Language Processing}, pages 6034--6055, 2022{\natexlab{b}}.

\bibitem[Singh et~al.(2024)Singh, Ravfogel, Herzig, Aharoni, Cotterell, and Kumaraguru]{singh2024representation}
S.~Singh, S.~Ravfogel, J.~Herzig, R.~Aharoni, R.~Cotterell, and P.~Kumaraguru.
\newblock Representation surgery: theory and practice of affine steering.
\newblock In \emph{Proceedings of the 41st International Conference on Machine Learning}, pages 45663--45680, 2024.

\bibitem[Vargas and Cotterell(2020)]{vargas2020exploring}
F.~Vargas and R.~Cotterell.
\newblock Exploring the linear subspace hypothesis in gender bias mitigation.
\newblock In B.~Webber, T.~Cohn, Y.~He, and Y.~Liu, editors, \emph{Proceedings of the 2020 Conference on Empirical Methods in Natural Language Processing (EMNLP)}, pages 2902--2913, Online, Nov. 2020. Association for Computational Linguistics.
\newblock \doi{10.18653/v1/2020.emnlp-main.232}.

\bibitem[Veitch et~al.(2021)Veitch, D'Amour, Yadlowsky, and Eisenstein]{veitch2021counterfactual}
V.~Veitch, A.~D'Amour, S.~Yadlowsky, and J.~Eisenstein.
\newblock Counterfactual invariance to spurious correlations in text classification.
\newblock \emph{Advances in neural information processing systems}, 34:\penalty0 16196--16208, 2021.

\bibitem[Voita and Titov(2020)]{voita2020information}
E.~Voita and I.~Titov.
\newblock Information-theoretic probing with minimum description length.
\newblock In B.~Webber, T.~Cohn, Y.~He, and Y.~Liu, editors, \emph{Proceedings of the 2020 Conference on Empirical Methods in Natural Language Processing (EMNLP)}, pages 183--196, Online, Nov. 2020. Association for Computational Linguistics.
\newblock \doi{10.18653/v1/2020.emnlp-main.14}.

\bibitem[Zhao and Gordon(2022)]{zhao2022inherent}
H.~Zhao and G.~J. Gordon.
\newblock Inherent tradeoffs in learning fair representations.
\newblock \emph{Journal of Machine Learning Research}, 23\penalty0 (57):\penalty0 1--26, 2022.

\end{thebibliography}

\appendix
\onecolumn

\section*{Supplementary material}
\vspace{2em}
\section{Impact Statement and Ethical Considerations}

This work explores the removal of demographic attributes information from pre-trained representations, with potential applications in fairness. However, we caution that the method should not be seen as a definitive solution to bias in neural models. Its effectiveness depends on context, including the data and fairness metrics in use. The experiments are limited to a specific set of datasets, which may not capture all forms of implicit gender bias, meaning some biases may persist post-application. Additionally, the method is tailored to a particular technical definition of bias and may not be robust to other forms of bias or adversarial approaches. We encourage future research to address these limitations and further investigate its applicability.

\section{Proof that $\mathcal{L}_p(U)$ is the distance to the closest projection to $UU^\top$}
\label{app:proof}

Let $\hat{P} = UU^\top$ and $P= U_rU_r^\top$ where $U_r$ $U_r$ is the matrix of eigenvectors corresponding to the non-zero eigenvalues of $\hat{P}$. 

$U_r^\top U_r = I_r$ and $P$ is an orthogonal projection.

Since $U_r$ is the matrix of eigenvectors corresponding to the non-zero eigenvalues of $\hat{P}$, we can express $\hat{P}$ as:

\[
\hat{P} = U_r \Lambda U_r^\top
\]

where $\Lambda$ is the diagonal matrix of eigenvalues of $\hat{P}$. Thus, we have:

\[
\hat{P} P = U_r \Lambda U_r^\top U_rU_r^\top = U_r \Lambda U_r^\top = \hat{P}
\]

Since both $\hat{P}$ and $ P $ are symmetric semi-positive definite matrices, we move to the Frobenius norm of the difference between $ \hat{P}$ and $ P $:

\[
\|\hat{P} - P \|_\mathrm{F}^2 = \mathrm{Tr}((\hat{P} - P)^2)
\]

Expanding the terms, we obtain:

\[
\|\hat{P} - P \|_\mathrm{F}^2 = \mathrm{Tr}(\hat{P}^2) + \mathrm{Tr}(P^2) - 2 \mathrm{Tr}(\hat{P} P)
\]

\textbf{Term 1: $ \mathrm{Tr}(\hat{P}^2)$}

Since $ \hat{P} = U U^{\top} $, we can use the property $ \| AB \|_\mathrm{F}^2 = \| BA \|_\mathrm{F}^2 $ to write:

\[
\mathrm{Tr}(\hat{P}^2) = \| U^\top U \|_\mathrm{F}^2
\]

\textbf{Term 2: $ \mathrm{Tr}(P^2) $}

Since $ P^2 = P $ (as $ P $ is an orthogonal projection matrix), and $ \text{rank}(P) = r $, we have:

\[
\mathrm{Tr}(P^2) = \mathrm{Tr}(P) = r = \| I_r \|_\mathrm{F}^2
\]

\textbf{Term 3: $ \mathrm{Tr}(\hat{P} P) $}

From the equation $ \hat{P} P = \hat{P} $, we conclude that:

\[
\mathrm{Tr}(\hat{P} P) = \mathrm{Tr}(\hat{P}) = \mathrm{Tr}(U^\top U)
\]

\textbf{Final Expression}

Now substituting these results into the expression for the Frobenius norm:

\[
\|\hat{P} - P \|_\mathrm{F}^2 = \| U^\top U \|_\mathrm{F}^2 + \| I_r \|_\mathrm{F}^2 - 2 \mathrm{Tr}(U^\top U)
\]

This simplifies to:

\[
\|\hat{P} - P \|_\mathrm{F}^2 = \| U^\top U - I_r \|_\mathrm{F}^2
\]

Finally, using Equation \ref{eq:loss_proj}:

\[
\|\hat{P} - P \|_\mathrm{F}^2 = \mathcal{L}_p(U)
\]

This completes the proof.

\newpage
\section{Training Details}
\label{app:training-details}
\subsection*{Hardware and Environment}
All experiments are conducted on an NVIDIA GeForce RTX 2080 Ti GPU with 11GB of VRAM and an Intel(R) Core(TM) i9-9900K CPU, using PyTorch 2.0.1 and Python 3.11.4. We employ CUDA 12.6 for accelerated computations.

\subsection*{Optimization and Hyperparameters for $\overline{\mathrm{\textbf{L}}}$EOPARD}
We use the Adam optimizer \cite{kingma2015adam} with the following hyperparameters:
\begin{itemize}
    \item Learning rate: $1 \times 10^{-3}$ for \textsc{GloVe} embeddings, $5 \times 10^{-4}$ for \textsc{Bias in Bios} and \textsc{DIAL}.
    \item Batch size: Full batch (10,777) for \textsc{GloVe} embeddings, batches of size 8192 for \textsc{Bias in Bios}, batches of size 2048 for \textsc{DIAL}.
    \item Weight decay: $0.0$ (to avoid interfering with the \textit{projection loss} (Eq. \ref{eq:loss_proj}).
\end{itemize}
Each training terminates after a fixed number of epochs ($N_e = 1000$ for \textsc{GloVe} embeddings, $N_e = 100$ for \textsc{Bias in Bios}, $N_e = 200$ for \textsc{DIAL}). To improve convergence, we apply a learning rate scheduler that reduces the learning rate by a factor of 10 after $\frac{N_e}{2}$ epochs.

\subsection*{Model-Specific Settings}
\begin{itemize}
    \item \textbf{$\overline{\mathrm{L}}$EOPARD}: $\gamma$ is scaled as $\frac{1}{r^2}$, using $\frac{200}{r^2}$ for \textsc{GloVe} embeddings and $\frac{100}{r^2}$ for \textsc{Bias in Bios} and Dial.
    \item \textbf{LEACE}: We relied on the publicly available implementation from \url{https://github.com/EleutherAI/concept-erasure}.
    \item \textbf{KRaM} and \textbf{FaRM}: We reimplemented KRaM \cite{basu2024robust} and FaRM \cite{basu2022learning}. We set $\lambda = 0.7$ in KRaM's loss. For \textsc{GloVe} and \textsc{Bias in Bios}, the erasure function results from training a 4-layer MLP. For Dial, the erasure map is a 7-layer MLP.
\end{itemize}

\subsection*{Evaluation Metrics}
Probing classifiers and downstream task models have been trained using \texttt{scikit-learn}'s MLPClassifier with the following settings:
\begin{itemize}
    \item Learning rate: $10^{-4}$ for the probes and the downstream classifiers.
    \item Max iterations: 20 for the probes and the downstream classifiers.  
    \item Batch size: 512 for the probes and the downstream classifiers.
\end{itemize}
All reported accuracy and fairness metrics are averaged over five independent runs to ensure robustness.
\\ \\
\noindent MDL is evaluated on the full training set for \textsc{GloVe} and \textsc{\textsc{DIAL}} and on a subset of 100k samples for \textsc{Bias in Bios}. 

\clearpage
\section{Supplementary results}
\label{app:supplementary-results}

\subsection{\textsc{GloVe}}

\begin{table*}[ht!]

\caption{Gender erasure in \textsc{GloVe} embeddings. $\ddagger$ denotes results reported from \cite{basu2022learning} and \cite{basu2024robust}.}

\label{tab:glove-results}

\centering

\begin{tabular}{llclccc}
\toprule
 & $a_z^{\mathrm{bin}}$ $\downarrow$ \scriptsize(\%) & MDL $\uparrow$ \scriptsize (kBits) & $a_z^{\mathrm{ter}}$ $\downarrow$ \scriptsize(\%) & $A_{50\%} \uparrow$ & WS-353 $\uparrow$ & Rank\\ 
\midrule
\textit{original}   & \textit{100.0\;\scriptsize$\pm$0.0} & \textit{0.1} & \textit{100.0\;\scriptsize$\pm$0.0} & \textit{1.00} & \textit{0.70} & \textit{300} \\
\textit{random}     & \textit{50.0} & -- & \textit{33.3} & \textit{0.50} & -- & \textit{300} \\
FaRM$^\ddagger$      & 53.9 & 24.6 & -- & 0.65 & -- & 247 \\
KRaM$^\ddagger$ & 52.6 & -- & -- &  0.65 & 0.48 & 246\vspace{0.5em} \\ 

LEACE & 96.3\;\scriptsize$\pm$0.1 & 3.5 & 81.1\;\scriptsize$\pm$0.5 & 0.94  & 0.69 & 298\vspace{0.5em}\\

FaRM & 52.7\;\scriptsize$\pm$0.4 & 25.3 & 36.4\;\scriptsize$\pm$0.6 &  0.67 & 0.57 & 295\\ 
FaRM cascaded & 54.0\;\scriptsize$\pm$0.6 & 25.0 & 37.2\;\scriptsize$\pm$0.8 &  0.66 & 0.56 & 297\\ 
KRaM & 53.7\;\scriptsize$\pm$0.4 & 27.9 & 36.1\;\scriptsize$\pm$0.5 &  0.62 & 0.58 & 289 \\ 
KRaM cascaded & 53.1\;\scriptsize$\pm$0.6 & 27.8 & 37.1\;\scriptsize$\pm$0.6 &  0.64 & 0.56 & 288\vspace{0.5em} \\

$\overline{\mathrm{L}}$EOPARD (standard)& & & & & & \\

\;\;$|$\;\;$r=275$ & 88.8\;\scriptsize$\pm$\;0.5 & 7.1 & 70.0\;\scriptsize$\pm$\;0.4 & 0.88  & 0.72 & 275 \\
\;\;$|$\;\;$r=250$ & 85.5\;\scriptsize$\pm$\;0.3 & 8.7 & 65.5\;\scriptsize$\pm$\;0.6 & 0.86  & 0.69 & 250 \\
\;\;$|$\;\;$r=225$ & 82.1\;\scriptsize$\pm$\;0.5 & 10.2 & 60.9\;\scriptsize$\pm$\;0.6 & 0.85 & 0.66 & 225 \\
\;\;$|$\;\;$r=200$ & 77.5\;\scriptsize$\pm$\;0.5 & 12.0 & 56.6\;\scriptsize$\pm$\;0.4 & 0.83  & 0.65 & 200 \\
\;\;$|$\;\;$r=175$ & 74.0\;\scriptsize$\pm$\;0.9 & 13.6 & 53.3\;\scriptsize$\pm$\;0.6 & 0.82  & 0.62 & 175 \\
\;\;$|$\;\;$r=150$ & 70.3\;\scriptsize$\pm$\;0.5 & 15.3 & 49.7\;\scriptsize$\pm$\;1.2 & 0.81  & 0.58 & 150 \\
\;\;$|$\;\;$r=125$ & 65.6\;\scriptsize$\pm$\;0.5 & 18.3 & 46.2\;\scriptsize$\pm$\;0.5 & 0.80  & 0.54 & 125 \\
\;\;$|$\;\;$r=100$ & 61.4\;\scriptsize$\pm$\;0.7 & 21.7 & 42.7\;\scriptsize$\pm$\;0.7 & 0.78  & 0.48 & 100 \\
\;\;$|$\;\;$r=75$ & 57.3\;\scriptsize$\pm$\;0.6 & 26.2 & 39.7\;\scriptsize$\pm$\;0.4 & 0.75  & 0.45 & 75 \\
\;\;$|$\;\;$r=50$ & 54.1\;\scriptsize$\pm$\;0.8 & 28.3 & 37.0\;\scriptsize$\pm$\;0.7 & 0.72  & 0.34 & 50 \\
\;\;$|$\;\;$r=25$ & 51.1\;\scriptsize$\pm$\;0.8 & 19.3 & 35.4\;\scriptsize$\pm$\;0.8 & 0.68  & 0.24 & 25\vspace{0.5em} \\

$\overline{\mathrm{L}}$EOPARD (cascaded)& & & & & & \\

\;\;$|$\;\;$r=275$ & 87.3\;\scriptsize$\pm$\;0.4 & 8.0 & 68.1\;\scriptsize$\pm$\;0.7 & 0.87  & 0.73 & 275 \\
\;\;$|$\;\;$r=250$ & 80.6\;\scriptsize$\pm$\;0.2 & 11.6 & 60.2\;\scriptsize$\pm$\;0.6 & 0.84  & 0.68 & 250 \\
\;\;$|$\;\;$r=225$ & 71.4\;\scriptsize$\pm$\;0.8 & 15.7 & 51.0\;\scriptsize$\pm$\;0.5 & 0.82 & 0.65 & 225 \\
\;\;$|$\;\;$r=200$ & 65.2\;\scriptsize$\pm$\;0.7 & 19.8 & 45.4\;\scriptsize$\pm$\;0.7 & 0.81  & 0.64 & 200 \\
\;\;$|$\;\;$r=175$ & 58.9\;\scriptsize$\pm$\;0.4 & 23.8 & 41.6\;\scriptsize$\pm$\;0.3 & 0.79  & 0.63 & 175 \\
\;\;$|$\;\;$r=150$ & 55.9\;\scriptsize$\pm$\;1.1 & 27.1 & 38.1\;\scriptsize$\pm$\;0.6 & 0.78  & 0.64 & 150 \\
\;\;$|$\;\;$r=125$ & 51.8\;\scriptsize$\pm$\;0.9 & 30.4 & 36.3\;\scriptsize$\pm$\;0.5 & 0.76  & 0.64 & 125 \\
\;\;$|$\;\;$r=100$ & 51.0\;\scriptsize$\pm$\;0.4 & 32.4 & 35.4\;\scriptsize$\pm$\;0.7 & 0.74  & 0.64 & 100 \\
\;\;$|$\;\;$r=75$ & 50.7\;\scriptsize$\pm$\;0.7 & 34.1 & 34.2\;\scriptsize$\pm$\;0.7 & 0.73  & 0.60 & 75 \\
\;\;$|$\;\;$r=50$ & 50.4\;\scriptsize$\pm$\;0.6 & 34.0 & 34.3\;\scriptsize$\pm$\;1.0 & 0.70  & 0.60 & 50 \\
\;\;$|$\;\;$r=25$ & 50.1\;\scriptsize$\pm$\;0.9 & 25.9 & 34.2\;\scriptsize$\pm$\;0.8 & 0.66  & 0.49 & 25 \\

\bottomrule
\end{tabular}

\end{table*}

\begin{figure}[ht!]
    \centering
    \includegraphics[angle=0,origin=c,width=0.45\linewidth]{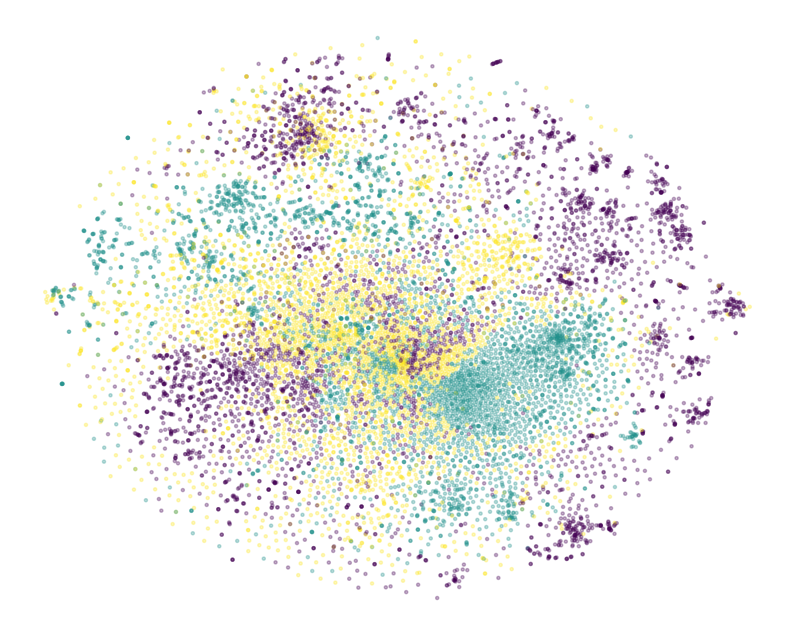}
    \includegraphics[angle=0,origin=c,width=0.45\linewidth]{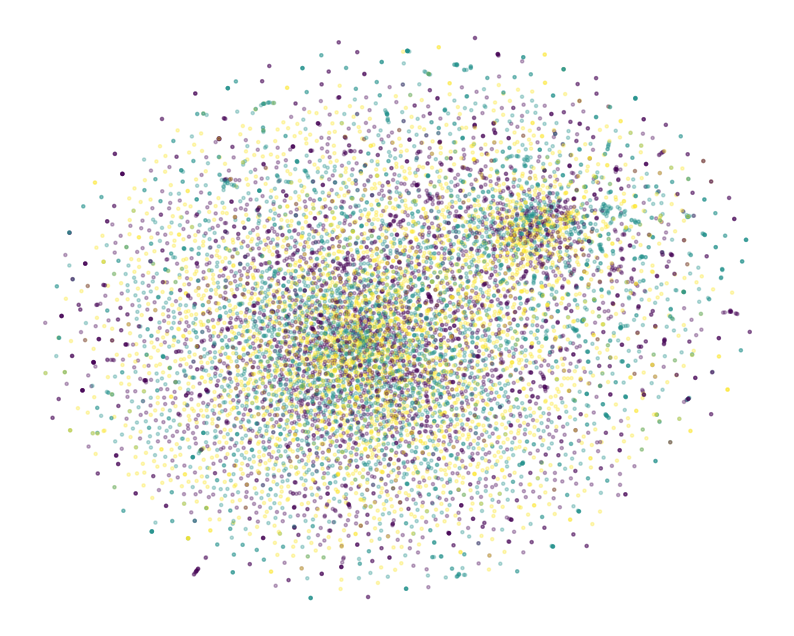}
    \caption{t-SNE visualization of \textsc{GloVe} embeddings before erasure (left) and after erasure via $\overline{\mathrm{L}}$EOPARD with a rank-150 orthogonal projection (right). Male-biased, female-biased, and neutral words are displayed in different colors.}
    \label{fig:app_tsne-glove}
\end{figure}

\clearpage
\subsection{\textsc{Bias in Bios}}

\begin{table}[h!]
\caption{Gender erasure in \textsc{Bias in Bios}. $\ddagger$ denotes results reported from \cite{basu2022learning} and \cite{basu2024robust}.}
\label{tab:biasbios-results}
\centering
% \resizebox{0.9\columnwidth}{!}{%

\begin{tabular}{llllc}
\toprule
 & $a_z$ $\downarrow$ \scriptsize(\%) & MDL & $a_y$ $\uparrow$ \scriptsize(\%)& Rank\\ 
\midrule
\textit{original}   & \textit{99.4\;\scriptsize $\pm$0.0} & \textit{2.7} & \textit{80.0\;\scriptsize$\pm$\;0.1}  & \textit{768} \\
\textit{random}     & \textit{53.5} & -- & \textit{33.5} & \textit{768} \\ \\
FaRM$^\ddagger$      & 55.6 & -- & 55.8 & -- \\ \\ 
LEACE & 98.6\;\scriptsize$\pm$\;0.1 & 15.7 & 79.8\;\scriptsize$\pm$\;0.2 & 767 \\ \\
FaRM & 58.8\;\scriptsize$\pm$\;0.2 & 236.9 &  55.6\;\scriptsize$\pm$\;0.1 & 759 \\
FaRM cascaded & 58.0\;\scriptsize$\pm$\;0.4 & 238.7 &  54.6\;\scriptsize$\pm$\;0.2 & --- \\
KRaM & 56.4\;\scriptsize$\pm$\;0.4 & 211.8 &  51.4\;\scriptsize$\pm$\;0.1 & 593\\ 
KRaM cascaded & 56.2\;\scriptsize$\pm$\;0.5 & 211.4 &  51.8\;\scriptsize$\pm$\;0.2 & ---\\ \\

$\overline{\mathrm{L}}$EOPARD& & & & \\

\;\;$|$\;\;$r=750$ & 98.4\;\scriptsize$\pm$\;0.1 & 31.0 & 79.6\;\scriptsize$\pm$\;0.2 & 750 \\
\;\;$|$\;\;$r=700$ & 97.0\;\scriptsize$\pm$\;0.9 & 55.7 & 79.5\;\scriptsize$\pm$\;0.2 & 700 \\
\;\;$|$\;\;$r=650$ & 96.7\;\scriptsize$\pm$\;0.4 & 77.4 & 79.4\;\scriptsize$\pm$\;0.2 & 650 \\
\;\;$|$\;\;$r=600$ & 95.6\;\scriptsize$\pm$\;0.2 & 90.9 & 79.1\;\scriptsize$\pm$\;0.3 & 600 \\
\;\;$|$\;\;$r=550$ & 94.3\;\scriptsize$\pm$\;1.0 & 97.5 & 79.0\;\scriptsize$\pm$\;0.2 & 550 \\
\;\;$|$\;\;$r=500$ & 93.4\;\scriptsize$\pm$\;0.8 & 112.0 & 78.9\;\scriptsize$\pm$\;0.1 & 500 \\
\;\;$|$\;\;$r=450$ & 92.8\;\scriptsize$\pm$\;0.6 & 110.7 & 78.6\;\scriptsize$\pm$\;0.4 & 450 \\
\;\;$|$\;\;$r=400$ & 91.6\;\scriptsize$\pm$\;0.8 & 119.1 & 78.4\;\scriptsize$\pm$\;0.3 & 400 \\
\;\;$|$\;\;$r=350$ & 90.1\;\scriptsize$\pm$\;0.8 & 126.4 & 78.2\;\scriptsize$\pm$\;0.1 & 350 \\
\;\;$|$\;\;$r=300$ & 88.9\;\scriptsize$\pm$\;0.3 & 128.3 & 77.9\;\scriptsize$\pm$\;0.1 & 300 \\
\;\;$|$\;\;$r=250$ & 88.2\;\scriptsize$\pm$\;0.6 & 133.8 & 77.4\;\scriptsize$\pm$\;0.2 & 250 \\
\;\;$|$\;\;$r=200$ & 85.7\;\scriptsize$\pm$\;0.4 & 142.3 & 76.8\;\scriptsize$\pm$\;0.1 & 200 \\
\;\;$|$\;\;$r=150$ & 82.1\;\scriptsize$\pm$\;0.4 & 158.7 & 75.5\;\scriptsize$\pm$\;0.2 & 150 \\
\;\;$|$\;\;$r=100$ & 77.1\;\scriptsize$\pm$\;0.7 & 172.0 & 73.2\;\scriptsize$\pm$\;0.2 & 100 \\
\;\;$|$\;\;$r=50$ & 69.0\;\scriptsize$\pm$\;0.7 & 186.2 & 67.9\;\scriptsize$\pm$\;0.2 & 50 \\
\;\;$|$\;\;$r=10$ & 58.8\;\scriptsize$\pm$\;0.7 & 193.4 & 53.4\;\scriptsize$\pm$\;0.1 & 10 \vspace{0.5em}\\

$\overline{\mathrm{L}}$EOPARD (cascaded)& & & & \\

\;\;$|$\;\;$r=750$ & 97.1\;\scriptsize$\pm$\;0.2 & 48.7 & 79.1\;\scriptsize$\pm$\;0.2 & 750 \\
\;\;$|$\;\;$r=700$ & 93.7\;\scriptsize$\pm$\;0.4 & 95.7 & 78.5\;\scriptsize$\pm$\;0.2 & 700 \\
\;\;$|$\;\;$r=650$ & 90.6\;\scriptsize$\pm$\;0.4 & 126.7 & 77.6\;\scriptsize$\pm$\;0.1 & 650 \\
\;\;$|$\;\;$r=600$ & 86.3\;\scriptsize$\pm$\;1.2 & 147.3 & 76.7\;\scriptsize$\pm$\;0.1 & 600 \\
\;\;$|$\;\;$r=550$ & 82.8\;\scriptsize$\pm$\;1.7 & 160.7 & 76.2\;\scriptsize$\pm$\;0.2 & 550 \\
\;\;$|$\;\;$r=500$ & 79.7\;\scriptsize$\pm$\;1.5 & 171.5 & 75.5\;\scriptsize$\pm$\;0.1 & 500 \\
\;\;$|$\;\;$r=450$ & 76.2\;\scriptsize$\pm$\;1.8 & 176.0 & 74.7\;\scriptsize$\pm$\;0.2 & 450 \\
\;\;$|$\;\;$r=400$ & 72.6\;\scriptsize$\pm$\;2.6 & 182.2 & 73.7\;\scriptsize$\pm$\;0.3 & 400 \\
\;\;$|$\;\;$r=350$ & 67.4\;\scriptsize$\pm$\;2.0 & 185.5 & 72.7\;\scriptsize$\pm$\;0.3 & 350 \\
\;\;$|$\;\;$r=300$ & 63.1\;\scriptsize$\pm$\;1.4 & 189.8 & 71.7\;\scriptsize$\pm$\;0.4 & 300 \\
\;\;$|$\;\;$r=250$ & 59.7\;\scriptsize$\pm$\;0.6 & 190.8 & 70.1\;\scriptsize$\pm$\;0.4 & 250 \\
\;\;$|$\;\;$r=200$ & 57.3\;\scriptsize$\pm$\;0.9 & 191.2 & 68.1\;\scriptsize$\pm$\;0.6 & 200 \\
\;\;$|$\;\;$r=150$ & 55.9\;\scriptsize$\pm$\;0.3 & 192.2 & 65.5\;\scriptsize$\pm$\;0.5 & 150 \\
\;\;$|$\;\;$r=100$ & 55.1\;\scriptsize$\pm$\;0.3 & 193.1 & 61.5\;\scriptsize$\pm$\;0.6 & 100 \\
\;\;$|$\;\;$r=50$ & 54.1\;\scriptsize$\pm$\;0.2 & 193.5 & 54.1\;\scriptsize$\pm$\;0.6 & 50 \\

\bottomrule
\end{tabular}
% }
\end{table}

\begin{figure}[ht!]
    \centering
    \includegraphics[angle=0,origin=c,width=0.45\linewidth]{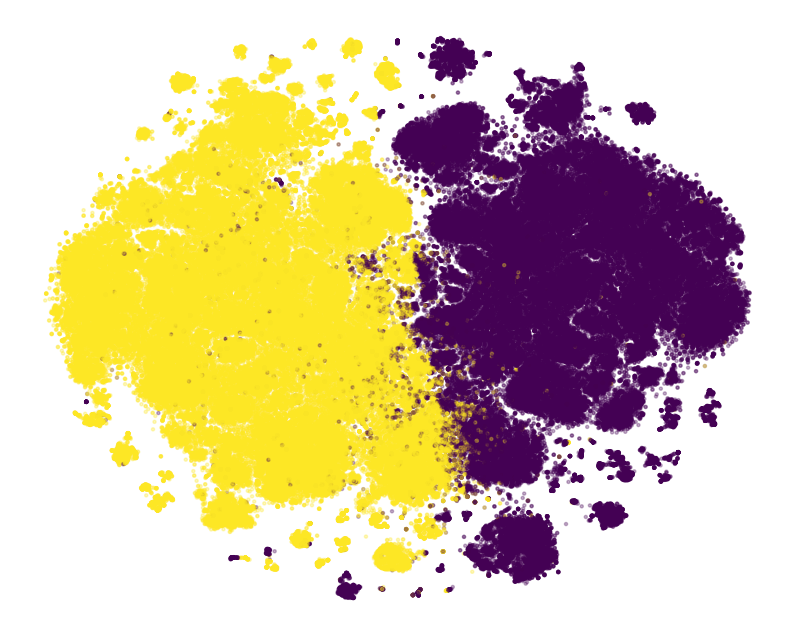}
    \includegraphics[angle=0,origin=c,width=0.45\linewidth]{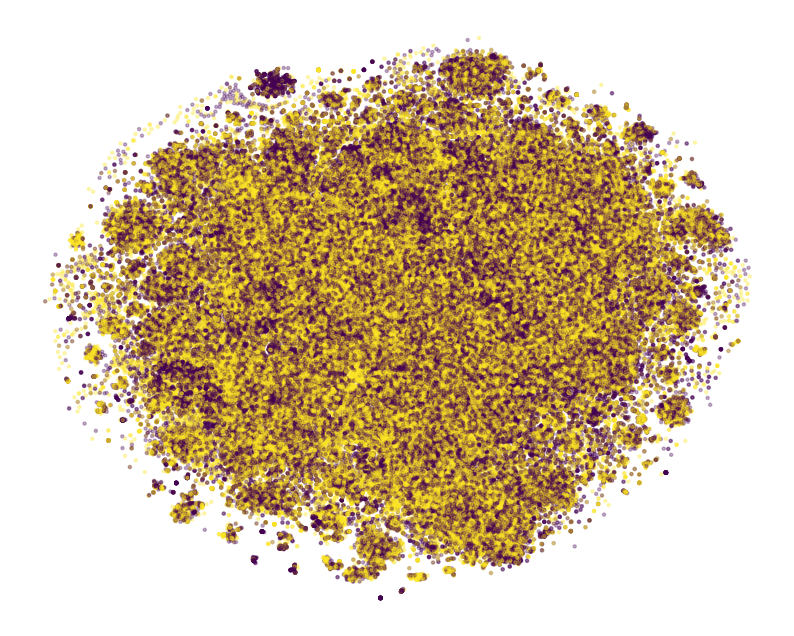}
    \caption{t-SNE visualization of \textsc{Bias in Bios} embeddings before erasure (left) and after erasure via cascaded $\overline{\mathrm{L}}$EOPARD with a rank-250 orthogonal projection (right). Female-biased and male-biased text representations are displayed in different colors.}
    \label{fig:app_tsne-biasbios}
\end{figure}

\clearpage
\subsection{\textsc{DIAL}}

\begin{table*}[ht!]
\caption{Race erasure in DIAL. $\ddagger$ denotes results reported from \cite{basu2022learning} and \cite{basu2024robust}.}
\label{tab:deepmoji-results}
\centering

\begin{tabular}{llllcr}
\toprule
 & $a_z$ $\downarrow$ & MDL $\uparrow$ &  $a_y$ $\uparrow$ & DP $\downarrow$ & rank\\ 
\midrule
\textit{original}   & \textit{88.0\;\scriptsize$\pm$\;0.1} & \textit{136.2} & \textit{75.8\;\scriptsize$\pm$\;0.1} & \textit{0.26} & \textit{300} \\
\textit{random}     & \textit{50.0} & & \textit{33.3} & -- & \textit{300}\\ 
FaRM$^\ddagger$   & 54.2 & -- & 74.8 & 0.09 & --\\
KRaM$^\ddagger$   & 54.0 & -- & 72.4 & 0.08 & --\\ \\

LEACE & 84.5\;\scriptsize$\pm$\;0.5 & 190.0 & 75.7\;\scriptsize$\pm$\;0.1 & 0.17 & 299\\ \\
FaRM & 53.7\;\scriptsize$\pm$\;0.8 & 335.8 & 73.9\;\scriptsize$\pm$\;0.3 & 0.06 & 264 \\
FaRM (cascaded) & 54.2\;\scriptsize$\pm$\;0.4 & 343.8 & 66.9\;\scriptsize$\pm$\;0.3 & 0.03 & 248\\
KRaM & 53.7\;\scriptsize$\pm$\;0.5 & 330.3 & 73.9\;\scriptsize$\pm$\;0.2 & 0.04 & 168\\
KRaM (cascaded) & 53.0\;\scriptsize$\pm$\;0.2 & 324.4 & 65.8\;\scriptsize$\pm$\;0.1 & 0.04 & 215\\ \\

$\overline{\mathrm{L}}$EOPARD & & & & & \\  

\;\;$|$\;\;$r=275$ & 86.3\;\scriptsize$\pm$\;0.2 & 176.0 &   75.5\;\scriptsize$\pm$\;0.2 & 0.19 & 275 \\
\;\;$|$\;\;$r=250$ & 86.0\;\scriptsize$\pm$\;0.3 & 183.1 &   75.3\;\scriptsize$\pm$\;0.2 & 0.21 & 250 \\
\;\;$|$\;\;$r=225$ & 85.4\;\scriptsize$\pm$\;0.2 & 193.1 &   75.3\;\scriptsize$\pm$\;0.1 & 0.21 & 225 \\
\;\;$|$\;\;$r=200$ & 84.8\;\scriptsize$\pm$\;0.2 & 200.6 &   75.2\;\scriptsize$\pm$\;0.2 & 0.18 & 200\\
\;\;$|$\;\;$r=175$ & 83.8\;\scriptsize$\pm$\;0.4 & 210.7 &   74.9\;\scriptsize$\pm$\;0.3 & 0.17 & 175\\
\;\;$|$\;\;$r=150$ & 82.8\;\scriptsize$\pm$\;0.2 & 223.4 &   74.9\;\scriptsize$\pm$\;0.1 & 0.17 & 150\\
\;\;$|$\;\;$r=125$ & 80.9\;\scriptsize$\pm$\;0.3 & 238.6 &   74.1\;\scriptsize$\pm$\;0.2 & 0.15 & 125\\
\;\;$|$\;\;$r=100$ & 78.9\;\scriptsize$\pm$\;0.5 & 248.3 &   74.4\;\scriptsize$\pm$\;0.1 & 0.18 & 100\\
\;\;$|$\;\;$r=75$ & 75.8\;\scriptsize$\pm$\;0.3 & 269.3 &   73.8\;\scriptsize$\pm$\;0.2 & 0.13 & 75\\
\;\;$|$\;\;$r=50$ & 70.2\;\scriptsize$\pm$\;0.7 & 290.7 &   71.6\;\scriptsize$\pm$\;0.2 & 0.13 & 50\\
\;\;$|$\;\;$r=40$ & 67.8\;\scriptsize$\pm$\;0.2 & 298.3 &   70.4\;\scriptsize$\pm$\;0.2 & 0.09 & 40\\
\;\;$|$\;\;$r=25$ & 61.1\;\scriptsize$\pm$\;0.4 & 307.9 &   68.4\;\scriptsize$\pm$\;0.1 & 0.04 & 25\\
\;\;$|$\;\;$r=15$ & 56.8\;\scriptsize$\pm$\;0.2 & 311.3 &   67.4\;\scriptsize$\pm$\;0.1 & 0.02 & 15\vspace{0.5em}\\

$\overline{\mathrm{L}}$EOPARD cascaded & & & & & \\  

\;\;$|$\;\;$r=275$ & 85.5\;\scriptsize$\pm$\;0.2 & 189.6 &   75.1\;\scriptsize$\pm$\;0.1 & 0.17 & 275 \\
\;\;$|$\;\;$r=250$ & 84.5\;\scriptsize$\pm$\;0.3 & 204.1 &   74.8\;\scriptsize$\pm$\;0.2 & 0.16 & 250\\
\;\;$|$\;\;$r=225$ & 83.8\;\scriptsize$\pm$\;0.3 & 213.7 &   74.8\;\scriptsize$\pm$\;0.1 & 0.15 & 225\\
\;\;$|$\;\;$r=200$ & 83.1\;\scriptsize$\pm$\;0.2 & 223.9 &   74.8\;\scriptsize$\pm$\;0.1 & 0.15 & 200\\
\;\;$|$\;\;$r=175$ & 81.3\;\scriptsize$\pm$\;0.2 & 236.0 &   74.6\;\scriptsize$\pm$\;0.3 & 0.13 & 175\\
\;\;$|$\;\;$r=150$ & 80.4\;\scriptsize$\pm$\;0.4 & 245.0 &   74.7\;\scriptsize$\pm$\;0.2 & 0.13 & 150 \\
\;\;$|$\;\;$r=125$ & 78.5\;\scriptsize$\pm$\;0.2 & 257.0 &   74.1\;\scriptsize$\pm$\;0.2 & 0.14 & 125\\
\;\;$|$\;\;$r=100$ & 77.4\;\scriptsize$\pm$\;0.3 & 264.1 &   74.4\;\scriptsize$\pm$\;0.2 & 0.14 & 100\\
\;\;$|$\;\;$r=75$ & 73.3\;\scriptsize$\pm$\;0.4 & 281.5 &   73.5\;\scriptsize$\pm$\;0.2 & 0.12 & 75\\
\;\;$|$\;\;$r=50$ & 67.6\;\scriptsize$\pm$\;0.3 & 297.6 &   71.8\;\scriptsize$\pm$\;0.2 & 0.08 & 50\\
\;\;$|$\;\;$r=40$ & 64.5\;\scriptsize$\pm$\;0.4 & 303.3 &   70.8\;\scriptsize$\pm$\;0.4 & 0.07 & 40\\
\;\;$|$\;\;$r=25$ & 62.4\;\scriptsize$\pm$\;0.3 & 303.0 &   72.8\;\scriptsize$\pm$\;0.2 & 0.11 & 25\\
\;\;$|$\;\;$r=15$ & 57.2\;\scriptsize$\pm$\;0.3 & 309.0 &   70.2\;\scriptsize$\pm$\;0.1 & 0.07 & 15\\
\bottomrule
\end{tabular}

\end{table*}

\begin{figure}[ht!]
    \centering
    \includegraphics[angle=0,origin=c,width=0.45\linewidth]{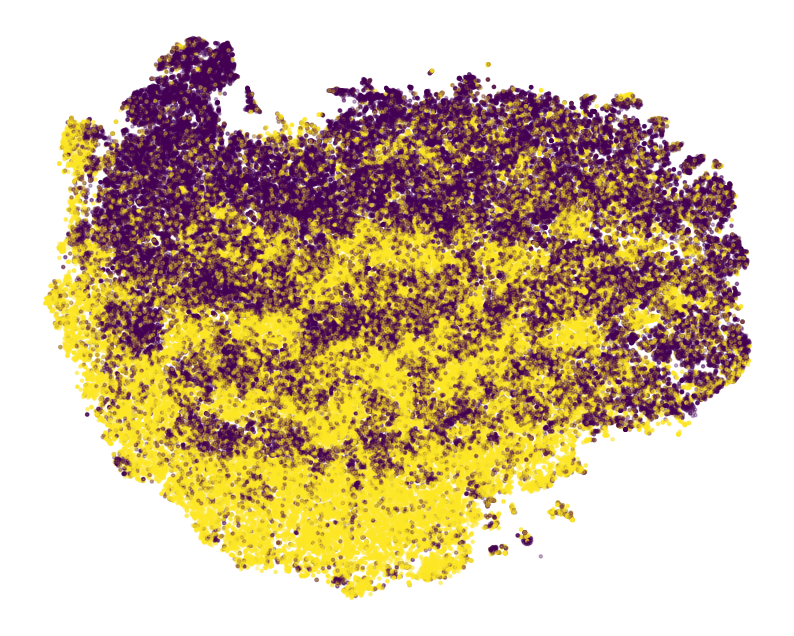}
    \includegraphics[angle=0,origin=c,width=0.45\linewidth]{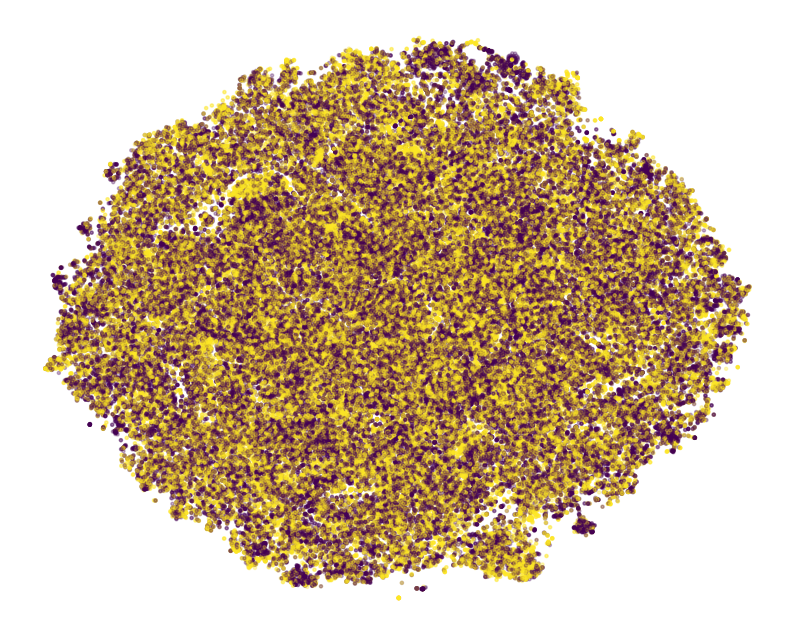}
    \caption{t-SNE visualization of \textsc{\textsc{DIAL}} embeddings before erasure (left) and after erasure via cascaded $\overline{\mathrm{L}}$EOPARD with a rank-50 orthogonal projection (right). Representations of tweets written in \emph{African-American English} and \emph{Standard American English} are displayed in different colors.}
    \label{fig:app_tsne-dial}
\end{figure}

\clearpage
\section{Batch size analysis}
\label{app:batch-size-analysis}

\subsection{\textsc{GloVe}}

\begin{figure}
\centering
\begin{minipage}{0.40\textwidth}
\centering
\captionsetup{type=table} %% tell latex to change to table

\caption{Gender erasure in \textsc{GloVe} with $\overline{\mathrm{L}}$EOPARD for different batch sizes used during training.}

\label{tab:app-glove-batch-analysis}

\centering
\resizebox{\linewidth}{!}{%
\begin{tabular}{lllclccc}
\toprule
 & rank & $a_z^{\mathrm{bin}}$ $\downarrow$ \scriptsize(\%) & MDL $\uparrow$ \scriptsize (kBits) & $a_z^{\mathrm{ter}}$ $\downarrow$ \scriptsize(\%) & $A_{50\%} \uparrow$ & WS-353 $\uparrow$ \\ 
\midrule

$\overline{\mathrm{L}}$EOPARD & & & & & & \\

\multirow{3}{10em}{\;\;$|$\;\;$r=275$} 
    & full batch & 88.8\;\scriptsize$\pm$\;0.5 & 7.1 & 70.0\;\scriptsize$\pm$\;0.4 & 0.88  & 0.72\\
    & 2048 & 88.0\;\scriptsize$\pm$\;0.2 & 7.7 & 68.9\;\scriptsize$\pm$\;0.9 & 0.88  & 0.73 \\
    & 512 & 87.5\;\scriptsize$\pm$\;0.1 & 7.7 & 68.2\;\scriptsize$\pm$\;0.7 & 0.88  & 0.71\vspace{0.5em} \\

\multirow{3}{10em}{\;\;$|$\;\;$r=250$} 
    & full batch & 85.5\;\scriptsize$\pm$\;0.3 & 8.7 & 65.5\;\scriptsize$\pm$\;0.6 & 0.86  & 0.69 \\
    & 2048 & 83.8\;\scriptsize$\pm$\;0.2 & 9.8 & 63.0\;\scriptsize$\pm$\;0.7 & 0.86  & 0.69 \\
    & 512 & 83.3\;\scriptsize$\pm$\;0.8 & 10.0 & 62.2\;\scriptsize$\pm$\;0.1 & 0.86  & 0.67\vspace{0.5em} \\
    
\multirow{3}{10em}{\;\;$|$\;\;$r=225$} 
    & full batch & 82.1\;\scriptsize$\pm$\;0.5 & 10.2 & 60.9\;\scriptsize$\pm$\;0.6 & 0.85 & 0.66 \\
    & 2048 & 78.5\;\scriptsize$\pm$\;0.8 & 11.5 & 58.1\;\scriptsize$\pm$\;0.6 & 0.84 & 0.66 \\
    & 512 & 78.3\;\scriptsize$\pm$\;0.5 & 11.7 & 57.2\;\scriptsize$\pm$\;0.4 & 0.84 & 0.65\vspace{0.5em} \\

\multirow{3}{10em}{\;\;$|$\;\;$r=200$} 
    & full batch & 77.5\;\scriptsize$\pm$\;0.5 & 12.0 & 56.6\;\scriptsize$\pm$\;0.4 & 0.83  & 0.65\\
    & 2048 & 74.3\;\scriptsize$\pm$\;0.3 & 13.5 & 54.3\;\scriptsize$\pm$\;0.6 & 0.83  & 0.63 \\
    & 512 & 74.8\;\scriptsize$\pm$\;0.4 & 13.4 & 54.0\;\scriptsize$\pm$\;0.7 & 0.83  & 0.63\vspace{0.5em} \\

\multirow{3}{10em}{\;\;$|$\;\;$r=175$} 
    & full batch & 74.0\;\scriptsize$\pm$\;0.9 & 13.6 & 53.3\;\scriptsize$\pm$\;0.6 & 0.82  & 0.62 \\
    & 2048 & 71.2\;\scriptsize$\pm$\;0.6 & 15.4 & 50.3\;\scriptsize$\pm$\;0.2 & 0.82  & 0.60 \\
    & 512 & 71.6\;\scriptsize$\pm$\;0.5 & 15.4 & 50.3\;\scriptsize$\pm$\;1.1 & 0.83  & 0.61\vspace{0.5em} \\
    
\multirow{3}{10em}{\;\;$|$\;\;$r=150$} 
    & full batch & 70.3\;\scriptsize$\pm$\;0.5 & 15.3 & 49.7\;\scriptsize$\pm$\;1.2 & 0.81  & 0.58 \\
    & 2048 & 66.6\;\scriptsize$\pm$\;0.4 & 17.7 & 47.1\;\scriptsize$\pm$\;0.4 & 0.81  & 0.57 \\
    & 512 & 68.1\;\scriptsize$\pm$\;0.7 & 17.0 & 47.8\;\scriptsize$\pm$\;0.7 & 0.82  & 0.58\vspace{0.5em} \\

\multirow{3}{10em}{\;\;$|$\;\;$r=125$} 
    & full batch & 65.6\;\scriptsize$\pm$\;0.5 & 18.3 & 46.2\;\scriptsize$\pm$\;0.5 & 0.80  & 0.54 \\
    & 2048 & 62.4\;\scriptsize$\pm$\;0.5 & 20.1 & 44.1\;\scriptsize$\pm$\;0.5 & 0.80  & 0.54 \\
    & 512 & 65.5\;\scriptsize$\pm$\;0.3 & 19.2 & 44.8\;\scriptsize$\pm$\;0.7 & 0.80  & 0.57\vspace{0.5em} \\

\multirow{3}{10em}{\;\;$|$\;\;$r=100$} 
    & full batch & 61.4\;\scriptsize$\pm$\;0.7 & 21.7 & 42.7\;\scriptsize$\pm$\;0.7 & 0.78  & 0.48 \\
    & 2048 & 59.5\;\scriptsize$\pm$\;0.8 & 23.4 & 40.0\;\scriptsize$\pm$\;0.6 & 0.78  & 0.47 \\
    & 512 & 61.2\;\scriptsize$\pm$\;0.9 & 22.8 & 42.3\;\scriptsize$\pm$\;0.7 & 0.79  & 0.52\vspace{0.5em} \\

\multirow{3}{10em}{\;\;$|$\;\;$r=75$} 
    & full batch & 57.3\;\scriptsize$\pm$\;0.6 & 26.2 & 39.7\;\scriptsize$\pm$\;0.4 & 0.75  & 0.45 \\
    & 2048 & 54.9\;\scriptsize$\pm$\;0.5 & 28.9 & 37.9\;\scriptsize$\pm$\;0.5 & 0.76  & 0.43 \\
    & 512 &  56.1\;\scriptsize$\pm$\;0.8 & 26.7 & 38.9\;\scriptsize$\pm$\;0.8 & 0.76  & 0.48\vspace{0.5em} \\

\multirow{3}{10em}{\;\;$|$\;\;$r=50$} 
    & full batch & 54.1\;\scriptsize$\pm$\;0.8 & 28.3 & 37.0\;\scriptsize$\pm$\;0.7 & 0.72  & 0.34 \\
    & 2048 & 52.5\;\scriptsize$\pm$\;1.2 & 30.1 & 35.7\;\scriptsize$\pm$\;0.3 & 0.72  & 0.36 \\
    & 512 & 53.1\;\scriptsize$\pm$\;0.5 & 30.1 & 36.4\;\scriptsize$\pm$\;0.4 & 0.73  & 0.37\vspace{0.5em} \\ 

\multirow{3}{10em}{\;\;$|$\;\;$r=25$} 
    & full batch & 51.1\;\scriptsize$\pm$\;0.8 & 19.3 & 35.4\;\scriptsize$\pm$\;0.8 & 0.68  & 0.24 \\
    & 2048 & 50.9\;\scriptsize$\pm$\;0.7 & 18.6 & 34.5\;\scriptsize$\pm$\;1.0 & 0.68  & 0.27 \\
    & 512 & 51.3\;\scriptsize$\pm$\;1.1 & 18.6 & 34.8\;\scriptsize$\pm$\;1.2 & 0.68  & 0.27\vspace{0.5em} \\

$\overline{\mathrm{L}}$EOPARD (cascaded)& & & & &  & \\

\multirow{3}{10em}{\;\;$|$\;\;$r=275$} 
    & full batch & 87.3\;\scriptsize$\pm$\;0.4 & 8.0 & 68.1\;\scriptsize$\pm$\;0.7 & 0.87  & 0.73 \\
    & 2048 & 87.7\;\scriptsize$\pm$\;0.5 & 7.8 & 68.5\;\scriptsize$\pm$\;0.5 & 0.87  & 0.73 \\
    & 512 & 87.7\;\scriptsize$\pm$\;0.3 & 7.8 & 68.6\;\scriptsize$\pm$\;0.2 & 0.87  & 0.74\vspace{0.5em} \\

\multirow{3}{10em}{\;\;$|$\;\;$r=250$} 
    & full batch & 80.6\;\scriptsize$\pm$\;0.2 & 11.6 & 60.2\;\scriptsize$\pm$\;0.6 & 0.84  & 0.68 \\
    & 2048 & 81.6\;\scriptsize$\pm$\;0.4 & 10.8 & 60.7\;\scriptsize$\pm$\;0.5 & 0.85  & 0.69 \\
    & 512 & 82.3\;\scriptsize$\pm$\;0.4 & 10.3 & 62.1\;\scriptsize$\pm$\;0.3 & 0.84  & 0.73\vspace{0.5em} \\
    
\multirow{3}{10em}{\;\;$|$\;\;$r=225$} 
    & full batch & 71.4\;\scriptsize$\pm$\;0.8 & 15.7 & 51.0\;\scriptsize$\pm$\;0.5 & 0.82 & 0.65 \\
    & 2048 & 74.6\;\scriptsize$\pm$\;0.7 & 13.8 & 53.6\;\scriptsize$\pm$\;0.9 & 0.83 & 0.68 \\
    & 512 & 77.4\;\scriptsize$\pm$\;0.2 & 12.7 & 56.2\;\scriptsize$\pm$\;0.6 & 0.82 & 0.71\vspace{0.5em}  \\

\multirow{3}{10em}{\;\;$|$\;\;$r=200$} 
    & full batch & 65.2\;\scriptsize$\pm$\;0.7 & 19.8 & 45.4\;\scriptsize$\pm$\;0.7 & 0.81  & 0.64\\
    & 2048 & 69.8\;\scriptsize$\pm$\;1.2 & 16.8 & 49.1\;\scriptsize$\pm$\;1.1 & 0.81  & 0.68 \\
    & 512 & 72.8\;\scriptsize$\pm$\;0.7 & 14.4 & 51.5\;\scriptsize$\pm$\;0.8 & 0.80  & 0.71\vspace{0.5em}  \\

\multirow{3}{10em}{\;\;$|$\;\;$r=175$} 
    & full batch & 58.9\;\scriptsize$\pm$\;0.4 & 23.8 & 41.6\;\scriptsize$\pm$\;0.3 & 0.79  & 0.63 \\
    & 2048 & 64.5\;\scriptsize$\pm$\;0.3 & 19.4 & 44.7\;\scriptsize$\pm$\;0.5 & 0.80  & 0.67 \\
    & 512 & 68.4\;\scriptsize$\pm$\;0.8 & 16.2 & 47.6\;\scriptsize$\pm$\;0.6 & 0.79  & 0.71\vspace{0.5em}  \\
    
\multirow{3}{10em}{\;\;$|$\;\;$r=150$} 
    & full batch & 55.9\;\scriptsize$\pm$\;1.1 & 27.1 & 38.1\;\scriptsize$\pm$\;0.6 & 0.78  & 0.64\\
    & 2048 & 60.5\;\scriptsize$\pm$\;0.6 & 22.7 & 42.1\;\scriptsize$\pm$\;0.5 & 0.78  & 0.66 \\
    & 512 & 65.7\;\scriptsize$\pm$\;0.4 & 18.4 & 45.5\;\scriptsize$\pm$\;0.6 & 0.78  & 0.69\vspace{0.5em}  \\

\multirow{3}{10em}{\;\;$|$\;\;$r=125$} 
    & full batch & 51.8\;\scriptsize$\pm$\;0.9 & 30.4 & 36.3\;\scriptsize$\pm$\;0.5 & 0.76  & 0.64 \\
    & 2048 & 57.2\;\scriptsize$\pm$\;0.6 & 26.1 & 39.3\;\scriptsize$\pm$\;1.0 & 0.77  & 0.65 \\
    & 512 & 62.5\;\scriptsize$\pm$\;0.8 & 20.6 & 43.4\;\scriptsize$\pm$\;0.9 & 0.77  & 0.67\vspace{0.5em}  \\

\multirow{3}{10em}{\;\;$|$\;\;$r=100$} 
    & full batch & 51.0\;\scriptsize$\pm$\;0.4 & 32.4 & 35.4\;\scriptsize$\pm$\;0.7 & 0.74  & 0.64 \\
    & 2048 & 54.5\;\scriptsize$\pm$\;0.2 & 29.1 & 37.1\;\scriptsize$\pm$\;0.3 & 0.75  & 0.63 \\
    & 512 & 59.8\;\scriptsize$\pm$\;1.0 & 24.4 & 40.8\;\scriptsize$\pm$\;0.5 & 0.76  & 0.63\vspace{0.5em}  \\

\multirow{3}{10em}{\;\;$|$\;\;$r=75$} 
    & full batch & 50.7\;\scriptsize$\pm$\;0.7 & 34.1 & 34.2\;\scriptsize$\pm$\;0.7 & 0.73  & 0.60 \\
    & 2048 & 53.1\;\scriptsize$\pm$\;0.7 & 33.0 & 35.4\;\scriptsize$\pm$\;0.9 & 0.74  & 0.60 \\
    & 512 & 57.0\;\scriptsize$\pm$\;0.6 & 29.0 & 38.0\;\scriptsize$\pm$\;0.6 & 0.74  & 0.58\vspace{0.5em}  \\

\multirow{3}{10em}{\;\;$|$\;\;$r=50$} 
    & full batch & 50.4\;\scriptsize$\pm$\;0.6 & 34.0 & 34.3\;\scriptsize$\pm$\;1.0 & 0.70  & 0.60\\
    & 2048 & 51.2\;\scriptsize$\pm$\;0.3 & 33.7 & 34.5\;\scriptsize$\pm$\;0.9 & 0.71  & 0.59 \\
    & 512 & 53.5\;\scriptsize$\pm$\;0.6 & 32.2 & 36.2\;\scriptsize$\pm$\;0.7 & 0.72  & 0.52\vspace{0.5em}  \\

\multirow{3}{10em}{\;\;$|$\;\;$r=25$} 
    & full batch & 50.1\;\scriptsize$\pm$\;0.9 & 25.9 & 34.2\;\scriptsize$\pm$\;0.8 & 0.66  & 0.49 \\
    & 2048 & 50.8\;\scriptsize$\pm$\;0.7 & 25.3 & 34.4\;\scriptsize$\pm$\;0.6 & 0.66  & 0.49 \\
    & 512 & 52.2\;\scriptsize$\pm$\;0.8 & 25.2 & 35.2\;\scriptsize$\pm$\;0.5 & 0.67  & 0.44\vspace{0.5em}  \\

\bottomrule
\end{tabular}
}
\end{minipage}
\hspace{2em}
\begin{minipage}{0.4\textwidth}
\centering
\begin{subfigure}[t]{\textwidth}
    \includegraphics[width=\textwidth]{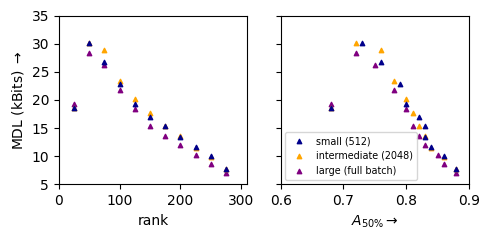}
    \caption{Standard training}
\end{subfigure}
\par\bigskip % force a bit of vertical whitespace
\begin{subfigure}[t]{\textwidth}
    \includegraphics[width=\textwidth]{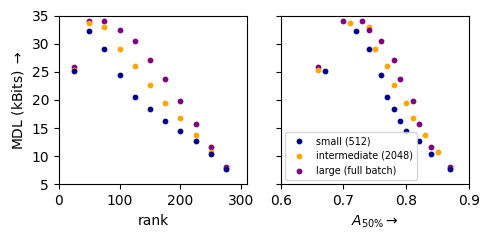}
    \caption{Cascaded training}
\end{subfigure}
\par\bigskip % force a bit of vertical whitespace
\caption{Gender erasure in \textsc{GloVe} with $\overline{\mathrm{L}}$EOPARD vs. rank of the feature space post-erasure (left) and Erasure–Utility trade-off (right) for different batch sizes used during training. The vertical axis, erasure strength as evaluated by MDL, is shared between the right and left panels.}
\label{fig:app-glove-batch-analysis}

\end{minipage}

\end{figure}

\clearpage

\subsection{\textsc{Bias in Bios}}

In order to limit computation costs, we have restricted the analysis to a reduced number of $r$ values.

\begin{figure}
\centering
\begin{minipage}{0.40\textwidth}
\centering
\captionsetup{type=table} %% tell latex to change to table
\caption{Gender erasure in \textsc{Bias in Bios} with $\overline{\mathrm{L}}$EOPARD for different batch sizes used during training.}
\label{tab:app-biasbios-batch-analysis}
\centering
\resizebox{\linewidth}{!}{%
\begin{tabular}{llllc}
\toprule
 & batch size & $a_z$ $\downarrow$ \scriptsize(\%) & MDL & $a_y$ $\uparrow$ \scriptsize(\%)\\ 
\midrule

$\overline{\mathrm{L}}$EOPARD& & & & \\

\multirow{3}{10em}{\;\;$|$\;\;$r=200$} 
    & 8192 & 85.7\;\scriptsize$\pm$\;0.4 & 142.3 & 76.8\;\scriptsize$\pm$\;0.1 \\
    & 2048 & 83.5\;\scriptsize$\pm$\;0.8 & 151.4 & 77.1\;\scriptsize$\pm$\;0.2\\
    & 512 & 83.7\;\scriptsize$\pm$\;0.4 & 146.0 & 77.5\;\scriptsize$\pm$\;0.2\vspace{0.5em}\\

\multirow{3}{10em}{\;\;$|$\;\;$r=150$} 
    & 8192 & 82.1\;\scriptsize$\pm$\;0.4 & 158.7 & 75.5\;\scriptsize$\pm$\;0.2 \\
    & 2048 & 80.0\;\scriptsize$\pm$\;0.4 & 164.5 & 76.3\;\scriptsize$\pm$\;0.1\\
    & 512 & 80.6\;\scriptsize$\pm$\;0.3 & 159.3 & 77.1\;\scriptsize$\pm$\;0.1\vspace{0.5em}\\

\multirow{3}{10em}{\;\;$|$\;\;$r=100$} 
    & 8192 & 77.1\;\scriptsize$\pm$\;0.7 & 172.0 & 73.2\;\scriptsize$\pm$\;0.2 \\
    & 2048 & 75.6\;\scriptsize$\pm$\;0.4 & 175.1 & 74.3\;\scriptsize$\pm$\;0.3\\
    & 512 & 76.3\;\scriptsize$\pm$\;0.5& 169.7 & 75.9\;\scriptsize$\pm$\;0.3\vspace{0.5em}\\

\multirow{3}{10em}{\;\;$|$\;\;$r=50$} 
    & 8192 & 69.0\;\scriptsize$\pm$\;0.7 & 186.2 & 67.9\;\scriptsize$\pm$\;0.2 \\
    & 2048 & 67.9\;\scriptsize$\pm$\;0.4 & 186.4 & 70.1\;\scriptsize$\pm$\;0.1\\
    & 512 & 69.9\;\scriptsize$\pm$\;0.2 & 183.8 & 72.6\;\scriptsize$\pm$\;0.2\vspace{0.5em}\\

$\overline{\mathrm{L}}$EOPARD (cascaded)& & & & \\

\multirow{3}{10em}{\;\;$|$\;\;$r=200$} 
    & 8192 & 57.3\;\scriptsize$\pm$\;0.9 & 191.2 & 68.1\;\scriptsize$\pm$\;0.6 \\
    & 2048 & 61.4\;\scriptsize$\pm$\;1.1 & 189.0 & 70.1\;\scriptsize$\pm$\;0.4\\
    & 512 & 65.3\;\scriptsize$\pm$\;0.0 & 187.7 & 70.7\;\scriptsize$\pm$\;0.4\vspace{0.5em}\\

\multirow{3}{10em}{\;\;$|$\;\;$r=150$} 
    & 8192 & 55.9\;\scriptsize$\pm$\;0.3 & 192.2 & 65.5\;\scriptsize$\pm$\;0.5 \\
    & 2048 & 59.8\;\scriptsize$\pm$\;1.1 & 189.9 & 68.1\;\scriptsize$\pm$\;0.3\\
    & 512 & 63.8\;\scriptsize$\pm$\;1.5 & 188.7 & 69.1\;\scriptsize$\pm$\;0.2\vspace{0.5em}\\

\multirow{3}{10em}{\;\;$|$\;\;$r=100$} 
    & 8192 & 55.1\;\scriptsize$\pm$\;0.3 & 193.1 & 61.5\;\scriptsize$\pm$\;0.6 \\
    & 2048 & 57.5\;\scriptsize$\pm$\;0.3 & 190.9 & 64.6\;\scriptsize$\pm$\;0.3\\
    & 512 & 62.1\;\scriptsize$\pm$\;0.5 & 187.9 & 66.7\;\scriptsize$\pm$\;0.3\vspace{0.5em}\\

\multirow{3}{10em}{\;\;$|$\;\;$r=50$} 
    & 8192 & 54.1\;\scriptsize$\pm$\;0.2 & 193.5 & 54.1\;\scriptsize$\pm$\;0.6 \\
    & 2048 & 54.9\;\scriptsize$\pm$\;0.4 & 192.2 & 58.3\;\scriptsize$\pm$\;0.4\\
    & 512 & 58.2\;\scriptsize$\pm$\;0.6 & 190.9 & 62.2\;\scriptsize$\pm$\;0.3\\

\bottomrule
\end{tabular}
}
\end{minipage}
\hspace{2em}
\begin{minipage}{0.4\textwidth}
\centering
\begin{subfigure}[t]{\textwidth}
    \includegraphics[width=\textwidth]{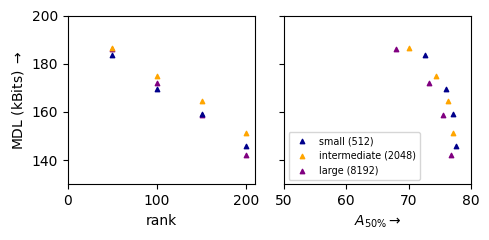}
    \caption{Standard training}
\end{subfigure}
\par\bigskip % force a bit of vertical whitespace
\begin{subfigure}[t]{\textwidth}
    \includegraphics[width=\textwidth]{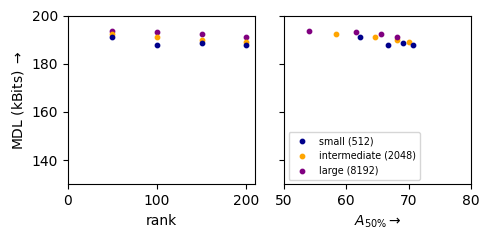}
    \caption{Cascaded training}
\end{subfigure}
\par\bigskip % force a bit of vertical whitespace
\caption{Gender erasure in \textsc{Bias in Bios} with $\overline{\mathrm{L}}$EOPARD vs. rank of the feature space post-erasure (left) and Erasure–Utility trade-off (right) for different batch sizes used during training. The vertical axis, erasure strength as evaluated by MDL, is shared between the right and left panels.}
\label{fig:app-biasbios-batch-analysis}

\end{minipage}

\end{figure}

\clearpage

\subsection{\textsc{DIAL}}

\begin{figure}
\centering
\begin{minipage}{0.40\textwidth}
\centering
\captionsetup{type=table} %% tell latex to change to table

\caption{Race erasure in \textsc{\textsc{DIAL}} with $\overline{\mathrm{L}}$EOPARD for different batch sizes used during training.}
\label{tab:app-deepmoji-batch-analysis}
\centering
\resizebox{\linewidth}{!}{%
\begin{tabular}{lcllcc}
\toprule
 & batch size & $a_z$ $\downarrow$ & MDL $\uparrow$ &  $a_y$ $\uparrow$ & DP $\downarrow$\\ 
\midrule

$\overline{\mathrm{L}}$EOPARD (standard) & & & & & \\

\multirow{3}{10em}{\;\;$|$\;\;$r=275$} 
        & 8192 & 86.5\;\scriptsize$\pm$\;0.2 & 172.7 &   75.7\;\scriptsize$\pm$\;0.1 & 0.19 \\
        & 2048 & 86.3\;\scriptsize$\pm$\;0.2 & 176.0 &   75.5\;\scriptsize$\pm$\;0.2 & 0.19 \\ 
        & 512  & 86.0\;\scriptsize$\pm$\;0.2 & 174.7 &   75.7\;\scriptsize$\pm$\;0.1 & 0.19 \vspace{0.5em}\\

\multirow{3}{10em}{\;\;$|$\;\;$r=250$} 
        & 8192 & 86.1\;\scriptsize$\pm$\;0.2 & 180.0 &   75.0\;\scriptsize$\pm$\;0.1 & 0.20 \\
        & 2048 & 86.0\;\scriptsize$\pm$\;0.3 & 183.1 &   75.3\;\scriptsize$\pm$\;0.2 & 0.21 \\
        & 512 & 85.7\;\scriptsize$\pm$\;0.3 & 180.6 &   75.4\;\scriptsize$\pm$\;0.2 & 0.18 \vspace{0.5em}\\

\multirow{3}{10em}{\;\;$|$\;\;$r=225$} 
        & 8192 & 85.8\;\scriptsize$\pm$\;0.1 & 188.5 &   74.8\;\scriptsize$\pm$\;0.2 & 0.19 \\
        & 2048 & 85.4\;\scriptsize$\pm$\;0.2 & 193.1 &   75.3\;\scriptsize$\pm$\;0.1 & 0.21 \\
        & 512 & 85.2\;\scriptsize$\pm$\;0.1 & 190.5 &   75.4\;\scriptsize$\pm$\;0.2 & 0.19 \vspace{0.5em}\\
        
\multirow{3}{10em}{\;\;$|$\;\;$r=200$} 
        & 8192 & 85.3\;\scriptsize$\pm$\;0.1 & 194.7 &   74.8\;\scriptsize$\pm$\;0.3 & 0.17 \\
        & 2048 & 84.8\;\scriptsize$\pm$\;0.2 & 200.6 &   75.2\;\scriptsize$\pm$\;0.2 & 0.18 \\
        & 512 & 84.4\;\scriptsize$\pm$\;0.2 & 197.1 &   75.3\;\scriptsize$\pm$\;0.2 & 0.18 \vspace{0.5em}\\

\multirow{3}{10em}{\;\;$|$\;\;$r=175$} 
        & 8192 & 84.7\;\scriptsize$\pm$\;0.3 & 205.0 &   74.6\;\scriptsize$\pm$\;0.3 & 0.16 \\
        & 2048 & 83.8\;\scriptsize$\pm$\;0.4 & 210.7 &   74.9\;\scriptsize$\pm$\;0.3 & 0.17 \\
        & 512 & 83.5\;\scriptsize$\pm$\;0.2 & 206.8 &   75.2\;\scriptsize$\pm$\;0.2 & 0.17 \vspace{0.5em}\\

\multirow{3}{10em}{\;\;$|$\;\;$r=150$} 
        & 8192 & 83.7\;\scriptsize$\pm$\;0.2 & 217.4 &   74.7\;\scriptsize$\pm$\;0.1 & 0.13 \\
        & 2048 & 82.8\;\scriptsize$\pm$\;0.2 & 223.4 &   74.9\;\scriptsize$\pm$\;0.1 & 0.17 \\
        & 512 & 82.2\;\scriptsize$\pm$\;0.3 & 217.7 &   75.0\;\scriptsize$\pm$\;0.2 & 0.17 \vspace{0.5em}\\

\multirow{3}{10em}{\;\;$|$\;\;$r=125$} 
        & 8192 & 82.3\;\scriptsize$\pm$\;0.4 & 229.0 &  74.2\;\scriptsize$\pm$\;0.2 & 0.13 \\
        & 2048 & 80.9\;\scriptsize$\pm$\;0.3 & 238.6 &   74.1\;\scriptsize$\pm$\;0.2 & 0.15 \\
        & 512 & 80.9\;\scriptsize$\pm$\;0.5 & 231.3 &   74.6\;\scriptsize$\pm$\;0.3 & 0.16 \vspace{0.5em}\\

\multirow{3}{10em}{\;\;$|$\;\;$r=100$} 
        & 8192 & 80.3\;\scriptsize$\pm$\;0.3 & 242.1 &   73.1\;\scriptsize$\pm$\;0.4 & 0.14 \\
        & 2048 & 78.9\;\scriptsize$\pm$\;0.5 & 248.3 &   74.4\;\scriptsize$\pm$\;0.1 & 0.18 \\
        & 512 & 79.4\;\scriptsize$\pm$\;0.3 & 241.2 &   74.2\;\scriptsize$\pm$\;0.2 & 0.16 \vspace{0.5em}\\

\multirow{3}{10em}{\;\;$|$\;\;$r=75$} 
        & 8192 & 76.4\;\scriptsize$\pm$\;0.6 & 265.6 &   72.9\;\scriptsize$\pm$\;0.1 & 0.12 \\
        & 2048 & 75.8\;\scriptsize$\pm$\;0.3 & 269.3 &   73.8\;\scriptsize$\pm$\;0.2 & 0.13 \\
        & 512 & 76.5\;\scriptsize$\pm$\;0.3 & 259.1 &   74.0\;\scriptsize$\pm$\;0.1 & 0.13 \vspace{0.5em}\\

\multirow{3}{10em}{\;\;$|$\;\;$r=50$} 
        & 8192 & 69.9\;\scriptsize$\pm$\;0.5 & 288.8 &  71.1\;\scriptsize$\pm$\;0.1 & 0.12 \\
        & 2048 & 70.2\;\scriptsize$\pm$\;0.7 & 290.7 &   71.6\;\scriptsize$\pm$\;0.2 & 0.13 \\
        & 512 & 71.6\;\scriptsize$\pm$\;0.6 & 280.3 &   72.2\;\scriptsize$\pm$\;0.2 & 0.11 \vspace{0.5em}\\

\multirow{3}{10em}{\;\;$|$\;\;$r=25$} 
        & 8192 & 61.2\scriptsize$\pm$\;0.4 & 309.1 &   69.1\;\scriptsize$\pm$\;.04 & 0.07 \\
        & 2048 & 61.1\;\scriptsize$\pm$\;0.4 & 307.9 &   68.4\;\scriptsize$\pm$\;0.1 & 0.04 \\
        & 512 & 64.6\;\scriptsize$\pm$\;0.3 & 300.5 &   70.1\;\scriptsize$\pm$\;0.1 & 0.10 \vspace{0.5em} \\

$\overline{\mathrm{L}}$EOPARD (cascaded) & & & & & \\  

\multirow{3}{10em}{\;\;$|$\;\;$r=275$} 
        & 8192 & 85.5\;\scriptsize$\pm$\;0.2 & 185.3 &   74.8\;\scriptsize$\pm$\;0.1 & 0.20 \\
        & 2048 & 85.5\;\scriptsize$\pm$\;0.2 & 189.6 &   75.1\;\scriptsize$\pm$\;0.1 & 0.17 \\
        & 512  & 85.2\;\scriptsize$\pm$\;0.2 & 193.4 &   75.4\;\scriptsize$\pm$\;0.2 & 0.16 \vspace{0.5em}\\

\multirow{3}{10em}{\;\;$|$\;\;$r=250$} 
        & 8192 & 85.0\;\scriptsize$\pm$\;0.3 & 198.7 &   74.9\;\scriptsize$\pm$\;0.3 & 0.19 \\
        & 2048 & 84.5\;\scriptsize$\pm$\;0.3 & 204.1 &   74.8\;\scriptsize$\pm$\;0.2 & 0.16 \\
        & 512 & 84.6\;\scriptsize$\pm$\;0.3 & 205.5 &   75.1\;\scriptsize$\pm$\;0.1 & 0.14 \vspace{0.5em}\\

\multirow{3}{10em}{\;\;$|$\;\;$r=225$} 
        & 8192 & 84.2\;\scriptsize$\pm$\;0.3 & 206.0 &   74.5\;\scriptsize$\pm$\;0.1 & 0.20 \\
        & 2048 & 83.8\;\scriptsize$\pm$\;0.3 & 213.7 &   74.8\;\scriptsize$\pm$\;0.1 & 0.15 \\
        & 512 & 83.8\;\scriptsize$\pm$\;0.3 & 214.0 &   75.3\;\scriptsize$\pm$\;0.1 & 0.13 \vspace{0.5em}\\
        
\multirow{3}{10em}{\;\;$|$\;\;$r=200$} 
        & 8192 & 83.4\;\scriptsize$\pm$\;0.2 & 215.1 &  74.6\;\scriptsize$\pm$\;0.2 & 0.20 \\
        & 2048 & 83.1\;\scriptsize$\pm$\;0.2 & 223.9 &   74.8\;\scriptsize$\pm$\;0.1 & 0.15 \\
        & 512 & 82.9\;\scriptsize$\pm$\;0.2 & 222.6 &   75.0\;\scriptsize$\pm$\;0.2 & 0.12 \vspace{0.5em}\\

\multirow{3}{10em}{\;\;$|$\;\;$r=175$} 
        & 8192 & 82.3\;\scriptsize$\pm$\;0.5 & 225.2 &   74.0\;\scriptsize$\pm$\;0.2 & 0.19 \\
        & 2048 & 81.3\;\scriptsize$\pm$\;0.2 & 236.0 &   74.6\;\scriptsize$\pm$\;0.3 & 0.13 \\
        & 512 & 81.8\;\scriptsize$\pm$\;0.2 & 232.1 &   74.8\;\scriptsize$\pm$\;0.1 & 0.13 \vspace{0.5em}\\

\multirow{3}{10em}{\;\;$|$\;\;$r=150$} 
        & 8192 & 80.4\;\scriptsize$\pm$\;0.5 & 239.4 &  74.0\;\scriptsize$\pm$\;0.2 & 0.17 \\
        & 2048 & 80.4\;\scriptsize$\pm$\;0.4 & 245.0 &   74.7\;\scriptsize$\pm$\;0.2 & 0.13 \\
        & 512 & 80.1\;\scriptsize$\pm$\;0.3 & 240.6 &   74.4\;\scriptsize$\pm$\;0.2 & 0.14 \vspace{0.5em}\\

\multirow{3}{10em}{\;\;$|$\;\;$r=125$} 
        & 8192 & 79.1\;\scriptsize$\pm$\;0.3 & 250.0 & 73.8\;\scriptsize$\pm$\;0.2 & 0.16 \\
        & 2048 & 78.5\;\scriptsize$\pm$\;0.2 & 257.0 &   74.1\;\scriptsize$\pm$\;0.2 & 0.14 \\
        & 512 & 78.1\;\scriptsize$\pm$\;0.4 & 253.0 &   74.4\;\scriptsize$\pm$\;0.1 & 0.11 \vspace{0.5em}\\

\multirow{3}{10em}{\;\;$|$\;\;$r=100$} 
        & 8192 & 77.2\;\scriptsize$\pm$\;0.4 & 263.3 &  73.9\;\scriptsize$\pm$\;0.2 & 0.15 \\
        & 2048 & 77.4\;\scriptsize$\pm$\;0.3 & 264.1 &   74.4\;\scriptsize$\pm$\;0.2 & 0.14 \\
        & 512 & 76.5\;\scriptsize$\pm$\;0.4 & 262.9 &   74.3\;\scriptsize$\pm$\;0.2 & 0.11 \vspace{0.5em}\\

\multirow{3}{10em}{\;\;$|$\;\;$r=75$} 
        & 8192 & 73.8\;\scriptsize$\pm$\;0.3 & 276.6 &  73.3\;\scriptsize$\pm$\;0.1 & 0.12 \\
        & 2048 & 73.3\;\scriptsize$\pm$\;0.4 & 281.53 &   73.5\;\scriptsize$\pm$\;0.2 & 0.12 \\
        & 512 & 73.1\;\scriptsize$\pm$\;0.3 & 274.7 &   74.0\;\scriptsize$\pm$\;0.3 & 0.11 \vspace{0.5em}\\

\multirow{3}{10em}{\;\;$|$\;\;$r=50$} 
        & 8192 & 68.6\;\scriptsize$\pm$\;0.5 & 293.2 &   71.9\;\scriptsize$\pm$\;0.3 & 0.09 \\
        & 2048 & 67.6\;\scriptsize$\pm$\;0.3 & 297.6 &   71.8\;\scriptsize$\pm$\;0.2 & 0.08 \\
        & 512 & 68.6\;\scriptsize$\pm$\;0.4 & 288.7 &   71.8\;\scriptsize$\pm$\;0.2 & 0.08 \vspace{0.5em}\\
        
\multirow{3}{10em}{\;\;$|$\;\;$r=25$} 
        & 8192 & 60.3\;\scriptsize$\pm$\;0.4 & 308.6 &  68.3\;\scriptsize$\pm$\;0.2 & 0.04 \\
        & 2048 & 62.4\;\scriptsize$\pm$\;0.3 & 303.0 &   72.8\;\scriptsize$\pm$\;0.2 & 0.11 \\
        & 512 & 61.8\;\scriptsize$\pm$\;0.5 & 304.5 &   69.0\;\scriptsize$\pm$\;0.1 & 0.07 \vspace{0.5em}\\

\bottomrule
\end{tabular}
}
\end{minipage}
\hspace{2em}
\begin{minipage}{0.4\textwidth}
\centering
\begin{subfigure}[t]{\textwidth}
    \includegraphics[width=\textwidth]{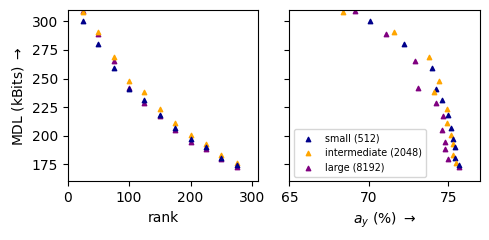}
    \caption{Standard training}
\end{subfigure}
\par\bigskip % force a bit of vertical whitespace
\begin{subfigure}[t]{\textwidth}
    \includegraphics[width=\textwidth]{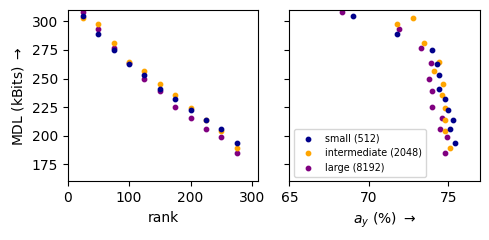}
    \caption{Cascaded training}
\end{subfigure}
\par\bigskip % force a bit of vertical whitespace
\caption{Race erasure in \textsc{\textsc{DIAL}} with $\overline{\mathrm{L}}$EOPARD vs. rank of the feature space post-erasure (left) and Erasure–Utility trade-off (right) for different batch sizes used during training. The vertical axis, erasure strength as evaluated by MDL, is shared between the right and left panels.}
\label{fig:app-dial-batch-analysis}
\end{minipage}

\end{figure}

\clearpage

\section{Mixture of kernels}
\label{app:kernel-analysis}

The analysis of results for various values of $M$, the number of kernels in the mixture of Eq.~\ref{eq:mixture-of-kernels}, shows that the improvement in performance becomes minimal beyond $M=2$.

\subsection{\textsc{GloVe}}

\begin{table}[ht!]
\centering
\caption{Gender erasure in \textsc{GloVe} with Cascaded $\overline{\mathrm{L}}$EOPARD for different number of kernels $M$ in the mixture.}

\label{tab:app-kernel-analysis-glove}

\centering
\resizebox{0.54\linewidth}{!}{%
\begin{tabular}{lllclccc}
\toprule
 rank $r$ & $M$ & $a_z^{\mathrm{bin}}$ $\downarrow$ \scriptsize(\%) & MDL $\uparrow$ \scriptsize (kBits) & $a_z^{\mathrm{ter}}$ $\downarrow$ \scriptsize(\%) & $A_{50\%} \uparrow$ & WS-353 $\uparrow$ \\ 
\midrule

\multirow{3}{3em}{200} 
    & 1 & 66.1\;\scriptsize$\pm$\;0.3 & 19.2 & 47.5\;\scriptsize$\pm$\;0.9 & 0.82 & 0.61 \\
    & 2 & 65.0\;\scriptsize$\pm$\;0.7 & 19.6 & 45.3\;\scriptsize$\pm$\;0.8 & 0.81 & 0.63 \\
    & 5 & 65.2\;\scriptsize$\pm$\;0.7 & 19.8 & 45.4\;\scriptsize$\pm$\;0.7 & 0.81 & 0.64 \vspace{0.5em} \\

\multirow{3}{3em}{150} 
    & 1 & 57.3\;\scriptsize$\pm$\;0.7 & 26.0 & 40.1\;\scriptsize$\pm$\;0.6 & 0.80 & 0.60 \\
    & 2 & 55.6\;\scriptsize$\pm$\;0.2 & 27.0 & 38.7\;\scriptsize$\pm$\;0.5 & 0.77  & 0.66 \\
    & 5 & 55.9\;\scriptsize$\pm$\;1.1 & 27.1 & 38.1\;\scriptsize$\pm$\;0.6 & 0.78  & 0.64\vspace{0.5em} \\
    
\multirow{3}{3em}{100} 
    & 1 & 53.5\;\scriptsize$\pm$\;0.7 & 31.9 & 36.1\;\scriptsize$\pm$\;1.4 & 0.77 & 0.57 \\
    & 2 & 51.2\;\scriptsize$\pm$\;0.4 & 32.5 & 35.0\;\scriptsize$\pm$\;0.7 & 0.74  & 0.65 \\
    & 5 & 51.0\;\scriptsize$\pm$\;0.4 & 32.4 & 35.4\;\scriptsize$\pm$\;0.7 & 0.74  & 0.64\vspace{0.5em} \\

\multirow{3}{3em}{50} 
    & 1 & 51.3\;\scriptsize$\pm$\;0.7 & 33.3 & 34.8\;\scriptsize$\pm$\;0.7 & 0.72  & 0.54 \\
    & 2 & 50.5\;\scriptsize$\pm$\;0.4 & 34.8 & 34.5\;\scriptsize$\pm$\;0.8 & 0.70  & 0.61 \\
    & 5 & 50.4\;\scriptsize$\pm$\;0.6 & 34.0 & 34.3\;\scriptsize$\pm$\;1.0 & 0.70  & 0.60\vspace{0.5em} \\

\bottomrule
\end{tabular}
}
\end{table}

\vspace{-1em}

\subsection{\textsc{Bias in Bios}}

\begin{table}[h!]
\caption{Gender erasure in \textsc{Bias in Bios} with Cascaded $\overline{\mathrm{L}}$EOPARD for different number of kernels $M$ in the mixture.}
\label{tab:app-kernel-analysis-biasbios}
\centering
\resizebox{0.34\linewidth}{!}{%

\begin{tabular}{llllc}
\toprule
 rank $r$ & $M$ & $a_z$ $\downarrow$ \scriptsize(\%) & MDL & $a_y$ $\uparrow$ \scriptsize(\%)\\ 
\midrule

\multirow{3}{3em}{200} 
    & 1 & 64.1\;\scriptsize$\pm$\;2.3 & 189.0 & 70.2\;\scriptsize$\pm$\;0.3 \\
    & 2 & 57.8\;\scriptsize$\pm$\;1.1 & 191.0 & 68.2\;\scriptsize$\pm$\;0.6  \\
    & 5 & 57.3\;\scriptsize$\pm$\;0.9 & 191.2 & 68.1\;\scriptsize$\pm$\;0.6  \\
    & 10 & 56.8\;\scriptsize$\pm$\;0.5 & 190.2 & 68.5\;\scriptsize$\pm$\;0.3 \vspace{0.5em}\\

\multirow{3}{3em}{150} 
    & 1 & 61.8\;\scriptsize$\pm$\;1.2 & 190.1 & 67.9\;\scriptsize$\pm$\;0.3 \\
    & 2 & 57.2\;\scriptsize$\pm$\;0.6 & 191.7 & 65.3\;\scriptsize$\pm$\;0.6  \\
    & 5 & 55.9\;\scriptsize$\pm$\;0.3 & 192.2 & 65.5\;\scriptsize$\pm$\;0.5  \\
    & 10 & 55.6\;\scriptsize$\pm$\;0.4 & 191.9 & 65.9\;\scriptsize$\pm$\;0.3 \vspace{0.5em}\\

\multirow{3}{3em}{100} 
    & 1 & 61.0\;\scriptsize$\pm$\;1.0 & 190.9 & 64.7\;\scriptsize$\pm$\;0.5 \\
    & 2 & 56.6\;\scriptsize$\pm$\;0.8 & 192.6 & 61.2\;\scriptsize$\pm$\;0.3  \\
    & 5 & 55.1\;\scriptsize$\pm$\;0.3 & 193.1 & 61.5\;\scriptsize$\pm$\;0.6   \\
    & 10 & 54.6\;\scriptsize$\pm$\;0.2 & 192.7 & 61.9\;\scriptsize$\pm$\;0.5 \vspace{0.5em}\\

\multirow{3}{3em}{50} 
    & 1 & 58.1\;\scriptsize$\pm$\;1.9 & 192.1 & 57.8\;\scriptsize$\pm$\;0.3 \\
    & 2 & 54.5\;\scriptsize$\pm$\;0.2 & 193.5 & 53.7\;\scriptsize$\pm$\;0.5  \\
    & 5 & 54.1\;\scriptsize$\pm$\;0.2 & 193.5 & 54.1\;\scriptsize$\pm$\;0.6  \\
    & 10 & 54.0\;\scriptsize$\pm$\;0.1 & 193.5 & 54.2\;\scriptsize$\pm$\;0.5 \vspace{0.5em}\\

\bottomrule
\end{tabular}
}
\end{table}
\vspace{-1em}

\subsection{\textsc{DIAL}}

\begin{table*}[h!]
\centering
\caption{Race erasure in \textsc{\textsc{DIAL}} with Cascaded $\overline{\mathrm{L}}$EOPARD for different number of kernels $M$ in the mixture.}
\label{tab:app-kernel-analysis-dial}
\centering
\resizebox{0.4\linewidth}{!}{%
\begin{tabular}{lcllcc}
\toprule
 rank $r$ & $M$ & $a_z$ $\downarrow$ & MDL $\uparrow$ &  $a_y$ $\uparrow$ & DP $\downarrow$\\ 
\midrule

\multirow{3}{3em}{75} 
        & 1 & 74.3\;\scriptsize$\pm$\;0.5 & 276.4 & 73.7\;\scriptsize$\pm$\;0.1 & 0.10 \\
        & 2 & 74.1\;\scriptsize$\pm$\;0.4 & 275.1 & 74.3\;\scriptsize$\pm$\;0.1 & 0.13 \\
        & 5 & 73.3\;\scriptsize$\pm$\;0.4 & 281.5 &   73.5\;\scriptsize$\pm$\;0.2 & 0.12 \\
        & 10 & 72.9\;\scriptsize$\pm$\;0.7 & 281.3 & 73.9\;\scriptsize$\pm$\;0.1 & 0.14 \vspace{0.5em}\\

\multirow{3}{3em}{50} 
        & 1 & 69.2\;\scriptsize$\pm$\;0.3 & 294.1 & 72.7\;\scriptsize$\pm$\;0.2 & 0.08 \\
        & 2 & 67.7\;\scriptsize$\pm$\;0.5 & 298.2 & 71.8\;\scriptsize$\pm$\;0.2 & 0.07 \\
        & 5 & 67.6\;\scriptsize$\pm$\;0.3 & 297.6 &   71.8\;\scriptsize$\pm$\;0.2 & 0.08  \\
        & 10 & 67.0\;\scriptsize$\pm$\;0.5 & 297.5 & 72.3\;\scriptsize$\pm$\;0.3 & 0.08 \vspace{0.5em}\\

\multirow{3}{3em}{25} 
        & 1 & 63.0\;\scriptsize$\pm$\;0.3 & 301.5 & 72.8\;\scriptsize$\pm$\;0.1 & 0.10 \\
        & 2 & 63.4\;\scriptsize$\pm$\;0.5 & 300.0 & 73.1\;\scriptsize$\pm$\;0.2 & 0.11 \\
        & 5 & 62.4\;\scriptsize$\pm$\;0.3 & 303.0 &   72.8\;\scriptsize$\pm$\;0.2 & 0.11  \\
        & 10 & 60.6\;\scriptsize$\pm$\;0.5 & 307.0 & 71.6\;\scriptsize$\pm$\;0.1 & 0.12 \vspace{0.5em}\\
        
\multirow{3}{3em}{15} 
        & 1 & 56.6\;\scriptsize$\pm$\;0.3 & 309.9 & 68.6\;\scriptsize$\pm$\;0.1 & 0.03  \\
        & 2 & 56.9\;\scriptsize$\pm$\;0.3 & 309.6 & 69.4\;\scriptsize$\pm$\;0.1 &  0.04\\
        & 5 & 57.2\;\scriptsize$\pm$\;0.3 & 309.0 &   70.2\;\scriptsize$\pm$\;0.1 & 0.07  \\
        & 10 & 54.2\;\scriptsize$\pm$\;0.2 & 311.6 & 68.6\;\scriptsize$\pm$\;0.1 & 0.07 \\

\bottomrule
\end{tabular}
}
\end{table*}

\end{document}